\documentclass[3p,12pt,times,number]{elsarticle}

\usepackage{amsmath,amssymb,mathtools} 
\usepackage{graphicx} 
\usepackage{booktabs,multirow,array,tabularx} 
\usepackage{url} 
\usepackage{xspace}
\usepackage{microtype} 
\usepackage{algorithm,algorithmic} 
\usepackage[table]{xcolor} 
\usepackage{caption} 
\usepackage{makecell} 
\usepackage{tcolorbox} 
\tcbuselibrary{listings,breakable,skins}
\usepackage[utf8]{inputenc}
\usepackage[T1]{fontenc}
\usepackage{textcomp}
\usepackage{subfig,amsfonts}
\usepackage{float} 
\usepackage{placeins} 
\usepackage[colorlinks,linkcolor=blue,citecolor=blue,urlcolor=blue]{hyperref}

\journal{Advanced Engineering Informatics}

\newtcblisting{prompt}{
  breakable,
  colback=gray!5,
  colframe=gray!50,
  left=2mm,right=2mm,top=1mm,bottom=1mm,
  boxrule=0.5pt,arc=1mm,
  listing only,
  listing options={
    basicstyle=\ttfamily\small,
    breaklines=true,
    columns=fullflexible,
    keepspaces=true,
    extendedchars=false,
  }
}

\newcommand{\cmark}{\checkmark}
\newcommand{\xmark}{\ensuremath{\times}}

\newcommand{\unit}[1]{\,\mathrm{#1}}

\begin{document}
\begin{frontmatter}

  \title{AeroTherm-GPT: A Verification-Centered LLM Framework for Thermal Protection System Engineering Workflows}
    \author[1]{QIAO Chuhan}

    \author[1]{ZHENG Jinglai}

    \author[2]{HUANG Jie}

    \author[1]{ZHAO Buyue}

    \author[1]{LI Fan}

    \author[1]{HUANG Haiming\corref{cor1}}
    \ead{huanghaiming@tsinghua.org.cn}

    \cortext[cor1]{Corresponding author.}

    \affiliation[1]{organization={School of Civil Engineering, Beijing Jiaotong University},
        city={Beijing},
        postcode={100044},
        country={China}}

    \affiliation[2]{organization={Beijing Research Institute of Telemetry},
        city={Beijing},
        postcode={100076},
        country={China}}
        
\begin{abstract}
Integrating Large Language Models (LLMs) into hypersonic thermal protection system (TPS) design is bottlenecked by cascading constraint violations when generating executable simulation artifacts. General-purpose LLMs, treating generation as single-pass text completion, fail to satisfy the sequential, multi-gate constraints inherent in safety-critical engineering workflows. To address this, we propose \textbf{AeroTherm-GPT}, the first TPS-specialized LLM Agent, instantiated through a \textbf{Constraint-Closed-Loop Generation (CCLG)} framework. CCLG organizes TPS artifact generation as an iterative workflow comprising generation, validation, CDG-guided repair, execution, and audit. The \textbf{Constraint Dependency Graph (CDG)} encodes empirical co-resolution structure among constraint categories, directing repair toward upstream fault candidates based on lifecycle ordering priors and empirical co-resolution probabilities. This upstream-priority mechanism resolves multiple downstream violations per action, achieving a Root-Cause Fix Efficiency of 4.16 versus 1.76 for flat-checklist repair. Evaluated on \textbf{HyTPS-Bench} and validated against external benchmarks, AeroTherm-GPT achieves 88.7\% End-to-End Success Rate (95\% CI: 87.5\textendash 89.9), a gain of +12.5\,pp over the matched non-CDG ablation baseline, without catastrophic forgetting on scientific reasoning and code generation tasks.
\end{abstract}
  
\begin{keyword}
Large Language Models \sep Hypersonic Thermal Protection Systems \sep Supervised Fine-Tuning \sep Engineering Problem Solving \sep Simulation Code Generation \sep Engineering Informatics
\end{keyword}
\end{frontmatter}

\section{Introduction}
Hypersonic Thermal Protection System (TPS) design requires synthesizing scientific literature, managing precise material properties \cite{poovathingal2017finite}, and executing high-fidelity numerical simulations under extreme aero-thermal environments \cite{candler2012nonequilibrium,anderson2006hypersonic,bertin1994hypersonic}. TPS components must withstand extreme heat fluxes through a combination of passive insulation, active cooling strategies such as opposing jets and transpiration cooling \cite{hu2025opposing,wu2023opposing}, and ablative materials including carbon/phenolic composites \cite{mansour2009carbon} and C/SiC composites \cite{zhu2014oxidation}. Accurate characterization of material thermal properties under high-temperature conditions, including temperature-dependent thermal conductivity and specific heat capacity \cite{schroeder2018inverse}, is essential for reliable simulation. Despite recent AI progress in data-driven surrogates and physics-informed learning, a critical bottleneck persists: converting heterogeneous requirements into runnable simulation code packages that satisfy strict unit, physics, and numerical constraints.

General-purpose LLMs fail to reliably close this gap because they treat generation as a one-pass text completion problem, merely approximating expert text rather than passing verifiable engineering gates \cite{raissi2019physics,ma2024llm4cfd}. In reality, engineering workflows require strict adherence to sequential and causal constraints. For instance, a simple unit mismatch can artificially inflate thermal diffusivity, destabilizing the numerical scheme and ultimately causing solver failure. Patching this downstream symptom without addressing the upstream cause traps standard LLMs in endless repair loops.

To address this cascading failure, this study proposes \textbf{AeroTherm-GPT}, a Constraint-Closed-Loop Generation (CCLG) Agent. Unlike standard LLMs, AeroTherm-GPT treats TPS design as a constrained search problem. We model the empirical dependency relations among engineering constraint categories using a \textbf{Constraint Dependency Graph (CDG)}, enabling the Agent to trace errors back to their root causes and prioritize upstream fixes.

The main contributions are:
\begin{enumerate}
\item The development of \textbf{AeroTherm-GPT}, the first TPS-specialized LLM Agent, which instantiates the CCLG framework.
\item The introduction of \textbf{CDG-guided repair}, which operationalizes upstream-priority error resolution based on empirical co-resolution statistics.
\item A dual-track evaluation protocol: \textbf{HyTPS-Bench} (a novel workflow-aligned benchmark) and external validation against external datasets.
\end{enumerate}

The remainder of this paper is organized as follows. Section~\ref{sec:related} reviews related work. Section~\ref{sec:aerotherm} presents the CCLG framework, CDG, and AeroTherm-GPT implementation. Section~\ref{sec:benchmark} describes the dual-track evaluation protocol and baselines. Section~\ref{sec:results} reports experimental results including ablation, failure analysis, and backbone transfer. Section~\ref{sec:case_studies} presents an end-to-end engineering workflow case study. Section~\ref{sec:discussion} discusses implications and limitations. Section~\ref{sec:conclusion} concludes the paper.

\begin{figure}
    \centering
    \includegraphics[height=24cm, width=\linewidth, keepaspectratio]{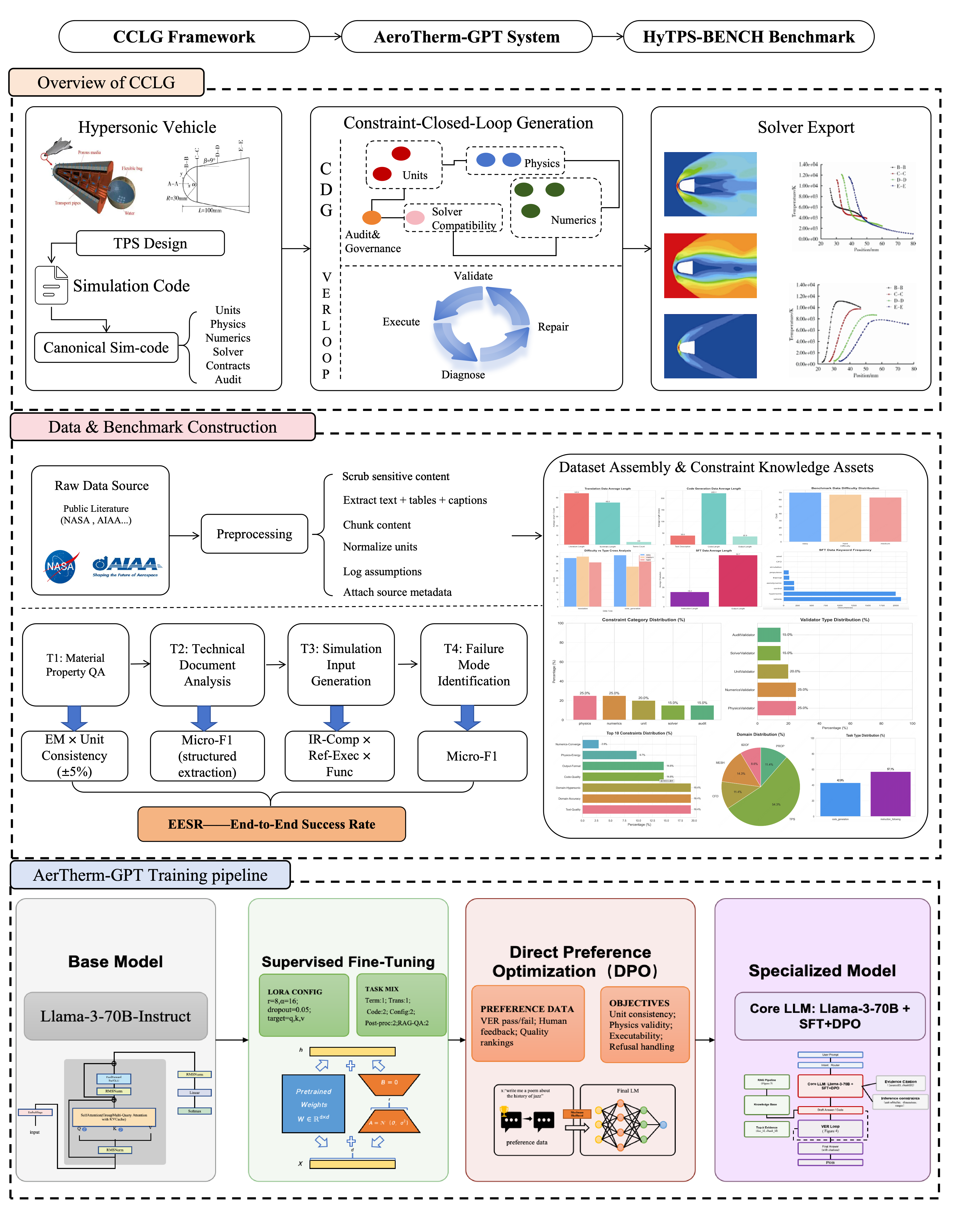}
    \caption{Overview of the proposed approach. An engineering requirement enters the CCLG workflow (generation $\to$ validation $\to$ CDG-guided repair $\to$ execution $\to$ audit), producing an executable and traceable simulation artifact. AeroTherm-GPT is a TPS-oriented instantiation of this framework; HyTPS-Bench provides the workflow-aligned evaluation setting.}
    \label{fig:overview}
\end{figure}

\section{Related Work}\label{sec:related}

\subsection{Engineering-Domain LLMs, Code Generation, and Verifiable Repair}
General-purpose LLMs have shown capabilities in scientific reasoning \cite{lightman2024verify,cobbe2021training,chen2023theoremqa}, and chain-of-thought prompting \cite{wei2022chain,kojima2022large} has further improved multi-step reasoning. Domain adaptation via fine-tuning and retrieval-augmented generation \cite{sun2024scieval} has improved performance on isolated tasks such as material property prediction \cite{shukla2024mlp} and technical QA \cite{rein2024gpqa}. In the hypersonic domain specifically, physics-informed neural networks (PINNs) have been applied to solve 2D non-Fourier heat conduction equations \cite{raissi2019physics}, predict thermal responses of charring ablative materials \cite{mortensen2015pyrolysis}, and simultaneously identify temperature-dependent thermal properties of TPS materials \cite{schroeder2018inverse}. Deep operator networks have demonstrated capability in solving compressible flow problems \cite{lu2021learning}, while surrogate-based optimization frameworks integrating PINNs with active learning and reinforcement learning have been proposed for engineering design tasks \cite{karniadakis2021physics}. LLM capability in hypersonic applications---covering fundamental knowledge, formula invocation, and automated programming---has been assessed as a domain benchmark \cite{zheng2025hyperkfa}. Recent work in engineering informatics has explored LLMs for knowledge graph construction in fault diagnosis \cite{liao2025llm}, structured maintenance document generation \cite{shi2025stepwise}, and integrating knowledge graphs with LLMs for industrial information systems \cite{wang2025cognitive}. Self-repair frameworks \cite{madaan2023self,chen2023teaching} and Reflexion-style approaches improve code generation robustness through iterative execution feedback, and tool-augmented agents \cite{schick2024toolformer,yao2023react} (e.g., SWE-Agent-style frameworks \cite{yang2024sweagent}) add retrieval and structured repair. However, these approaches either address individual capabilities in isolation or treat errors as separate symptoms. In engineering artifact generation, constraint violations propagate: a unit mismatch inflates diffusivity, triggering downstream numerical instability and execution failure. Patching the downstream symptom without resolving the upstream root cause causes the repair loop to cycle. CCLG addresses this by encoding empirical constraint propagation structure in the CDG, enabling upstream-priority repair (RCFE 4.16 vs.\ 1.76 for flat checklist; Table~\ref{tab:cdg_rcfe}) that resolves multiple downstream violations per action and treats the TPS workflow as a single constrained search problem with explicit multi-gate quality requirements.

\subsection{Engineering Informatics: Knowledge Representation, Traceability, and Benchmarks}
Engineering ontologies \cite{gruber1995toward} and design rationale capture \cite{lee1997design} have established that structured knowledge representation and decision traceability are critical in safety-critical domains---motivating CCLG's constraint asset schema and audit gate. Knowledge graph-based approaches have demonstrated the value of structured engineering knowledge for design reasoning and manufacturing process support \cite{haruna2024knowledge,pan2024unifying}. MBSE and digital twin paradigms \cite{grieves2017digital,madni2019leveraging,li2024digital} advocate lifecycle-aware simulation integration and ontology-based modelling, which CCLG operationalizes through explicit verification gates. Agentic frameworks with process reward models \cite{lightman2024verify,snell2025scaling,uesato2022solving} and verifiable reasoning \cite{wang2024respecting} improve multi-step reliability but do not encode the engineering lifecycle ordering (unit $\to$ physics $\to$ numerics $\to$ execution $\to$ audit) that governs safety-critical artifact generation.

The CDG is conceptually related to failure propagation chains in Failure Mode and Effects Analysis (FMEA), where cascading effects between failure modes are identified to guide corrective action priority. Standard FMEA encodes failure propagation qualitatively---typically through severity, occurrence, and detection ratings aggregated into a Risk Priority Number---and relies on domain experts to traverse the fault chain manually. The CDG operationalizes the same underlying principle---that upstream failures propagate to downstream symptoms, and that correcting the root cause is more efficient than patching symptoms---but quantifies propagation structure as empirical conditional probabilities calibrated from repair traces, enabling automated upstream-priority repair in a generative AI pipeline. This shifts failure propagation reasoning from a qualitative human-in-the-loop activity to a data-informed, machine-executable repair strategy.

Existing LLM benchmarks \cite{hendrycks2021measuring,wang2024scibench} evaluate knowledge or procedural reasoning but omit industrial KPIs such as execution success rates, constraint compliance, and audit completeness. HyTPS-Bench addresses this gap with workflow-aligned tasks and executable engineering metrics across the full TPS lifecycle.

\section{Verification-Centered Framework for TPS Engineering Workflows}\label{sec:aerotherm}

\subsection{Problem Formulation}\label{sec:formulation}

Let \(a\) denote a simulation code package (canonical specification + runnable solver and analysis code). The goal is to produce \(a\) satisfying a constraint set \(\mathcal{C}=\{c_i\}_{i=1}^M\) spanning units \cite{buckingham1914physically}, physics, numerics, execution contracts, and auditability. Each constraint \(c_i\) exposes an executable validator \(v_i(a)\in\{0,1\}\), and a package is ready if
\begin{equation}\label{eq:ready_predicate}
\mathrm{Ready}(a) = \prod_{i=1}^M v_i(a) = 1.
\end{equation}
Equation~\eqref{eq:ready_predicate} defines a \emph{completeness predicate}: $\mathrm{Ready}(a)=1$ requires all validators to simultaneously return 1, which is the criterion for accepting an artifact. This product form does \emph{not} imply that validators are evaluated in arbitrary order. Operationally, gates are evaluated sequentially (unit $\to$ physics $\to$ numerics $\to$ execution $\to$ audit) with short-circuit termination: if an upstream gate fails, downstream gates are not evaluated, since their results are physically undefined until upstream constraints are satisfied (e.g., the Fourier number stability condition is meaningless if the thermal conductivity unit is wrong). The product notation captures the acceptance criterion; the lifecycle ordering governs evaluation procedure. For transient TPS conduction, $\rho c_p \partial T/\partial t = \nabla\cdot(k\nabla T)$, the physics gates enforce $k,\rho,c_p>0$, and the numerical stability gate requires the Fourier number $\mathrm{Fo}=k\Delta t/(\rho c_p \Delta x^2)\le 1/2$, alongside unit gates on $k$, $\rho$, $c_p$, and $q''$. General-purpose LLMs fail this sequential structure because they treat generation as one-pass text completion rather than constrained search over artifact quality.

\begin{figure}[!t]
\centering
\includegraphics[width=1\textwidth]{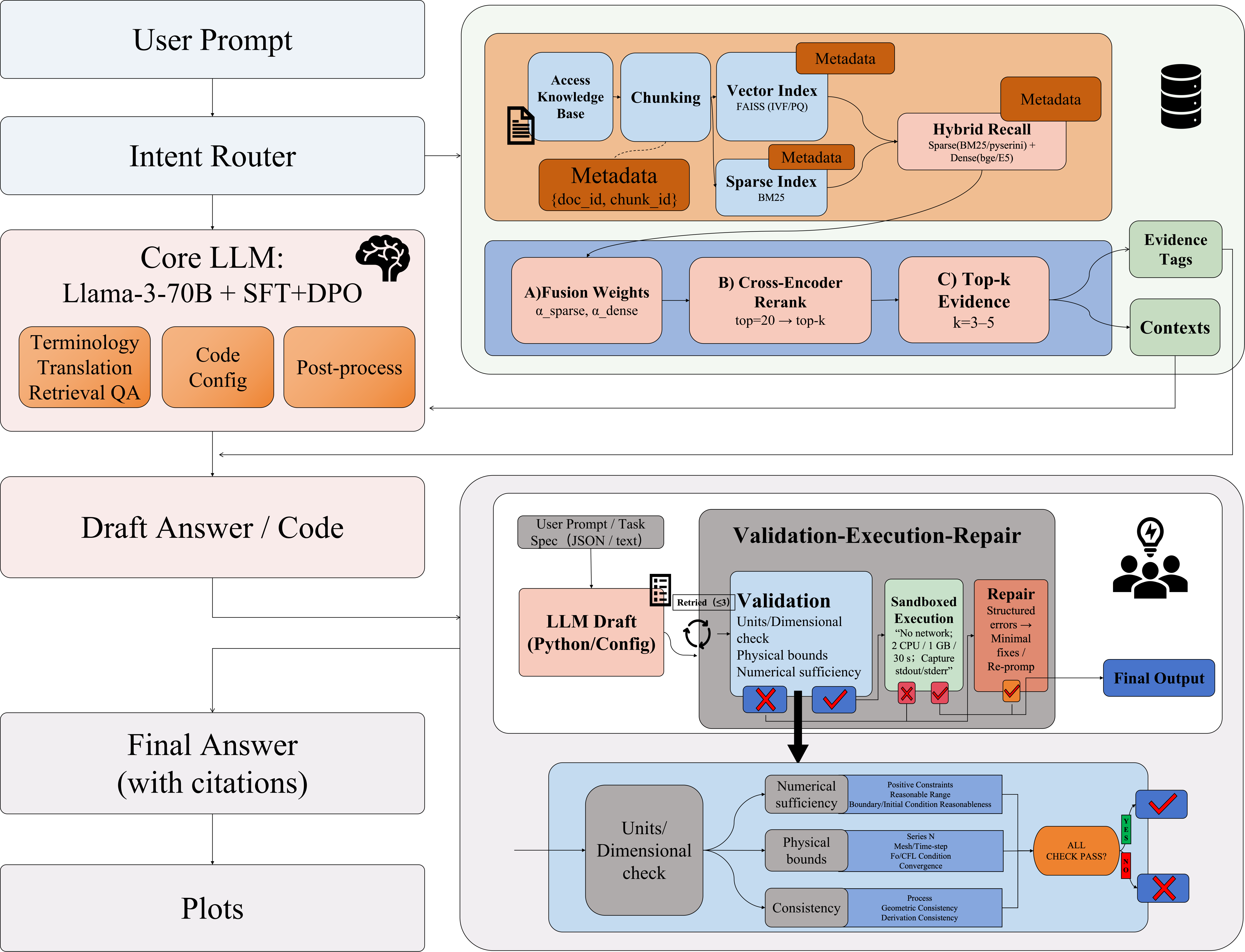}
\caption{End-to-end architecture of AeroTherm-GPT. Solid-line modules (constraint-aware RAG, TPS-specific SFT, CDG-aware VER loop, constraint-aware DPO, reference executor) are fully implemented and evaluated in all experiments. Dashed-line extensions (probabilistic constraint bounds, multi-fidelity CDG, edge-deployed variants) represent directions for future work and are not evaluated here.}
\label{fig:c2lg_architecture}
\end{figure}

\subsection{CCLG: Closed-Loop Generation, Validation, Repair, Execution, and Audit}

CCLG organizes TPS artifact generation as an iterative workflow rather than a one-pass completion. Given a natural-language engineering requirement, the framework proceeds through five stages: (1) \emph{generation}---producing a canonical specification and runnable code package; (2) \emph{validation}---checking each constraint gate \(v_i(a)\); (3) \emph{repair}---identifying and fixing violations, guided by the CDG; (4) \emph{execution}---running the code package in a sandboxed environment; and (5) \emph{audit}---recording provenance evidence linking every decision to cited sources, conversions, and validation outcomes. This loop continues until \(\mathrm{Ready}(a)=1\) or a compute budget is exhausted.

\subsubsection{Constraint Assets}
Each engineering constraint is represented as a structured knowledge asset with a source reference, applicability scope, severity level, a machine-executable validator, and repair guidance. The five constraint categories---unit, physical, numerical, execution-compatibility, and audit---are formalized as JSON-schema assets \cite{geng2023grammar} (see released constraint assets), enabling programmatic validation and targeted retrieval during repair.

\subsubsection{Violation-Triggered Retrieval and Audit}

\begin{figure}[!ht]
\centering
\includegraphics[width=0.95\linewidth]{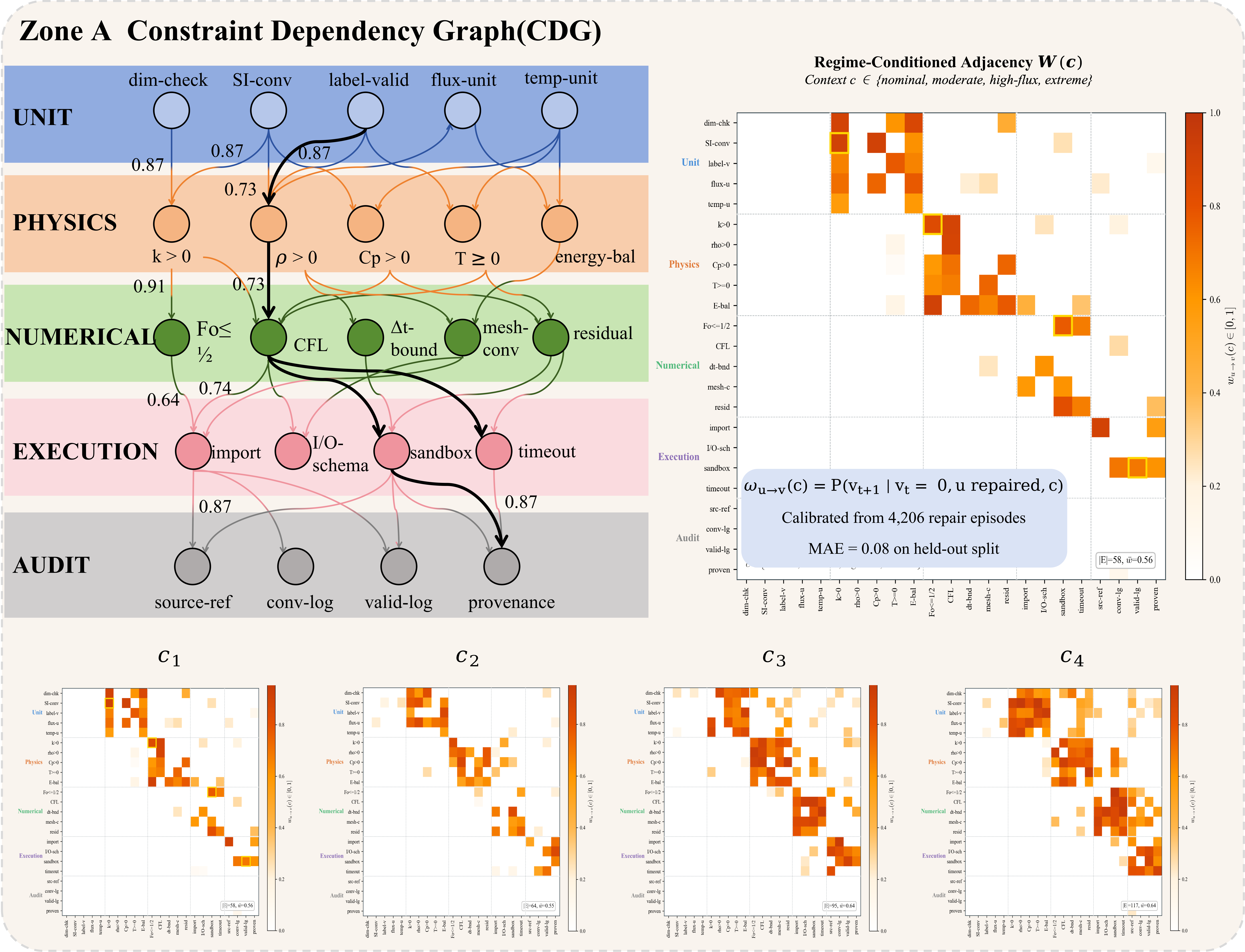} 
\caption{Constraint Dependency Graph (CDG) structure and regime-conditioned calibration. 
Top-left (Zone A): five-tier CDG with 23 nodes organized by lifecycle ordering (Unit $\rightarrow$ Physics $\rightarrow$ Numerical $\rightarrow$ Execution $\rightarrow$ Audit); directed edges represent empirical co-resolution probabilities calibrated from 4,206 repair episodes. 
Top-right: regime-conditioned adjacency matrix $W(c)$ across four thermal contexts (nominal, moderate, high-flux, extreme), with edge weights stratified by flight regime; mean absolute calibration error $= 0.08$ on 20\% held-out split ($n = 841$). 
Bottom: per-regime CDG visualizations ($c_1$--$c_4$) illustrating how edge weight distributions shift across operating conditions.
}
\label{fig:cdg_ver_loop_small}
\end{figure}

When a constraint violation is detected, CCLG triggers constraint-conditioned retrieval targeting the relevant standard, material property, or modeling rule. Retrieved documents carry provenance metadata (ID, page, confidence) recorded in the audit trail. The system also produces a structured design memo covering problem framing, decomposition, and regime conditions, and records every post-processing and validation step as provenance evidence.

\begin{figure}[!p]
\centering
\includegraphics[width=0.85\linewidth]{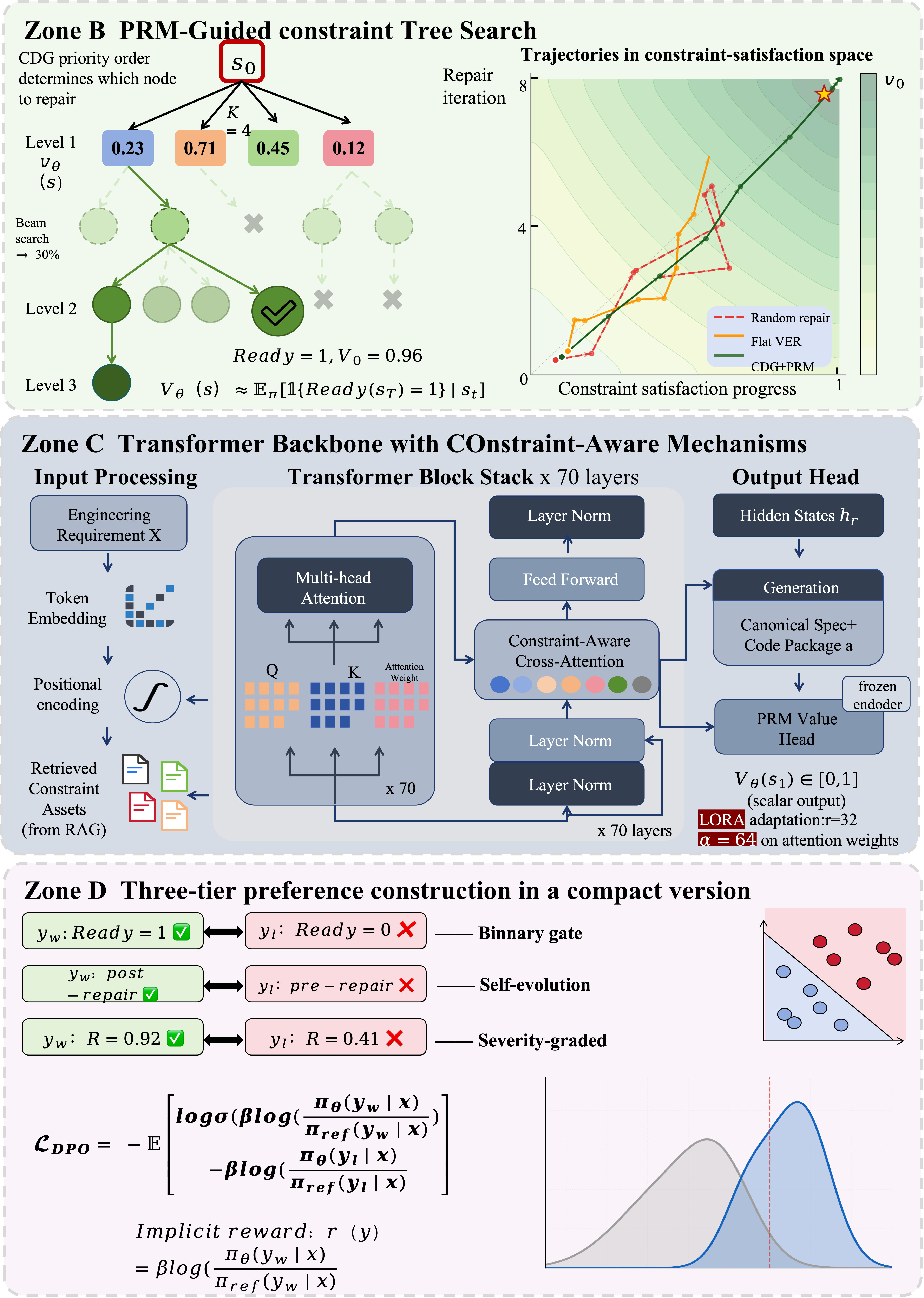}
\caption{Core computational mechanisms of AeroTherm-GPT. Zone B (top): PRM-guided constraint tree search (CDG priority, $K=4$ candidates scored by $V_\theta(s)$), terminating at $\text{Ready}(s)=1$ or budget exhaustion; right panel compares repair strategy trajectories. Zone C (middle): Transformer backbone with RAG cross-attention, frozen encoder, PRM head ($V_\theta(s)\in[0,1]$), and LoRA adaptation ($r=32, \alpha=64$). Zone D (bottom): Three-tier DPO preference construction (binary, self-evolution, severity-graded) with implicit reward distribution.}
\label{fig:cdg_ver_loop}
\end{figure}

\begin{table}[!t]
\centering
\scriptsize
\tabcolsep=4pt
\begin{tabular}{@{}lccc@{}}
\toprule
Repair Strategy & Root-Cause Fix Efficiency & Avg Iter. & Rework Red. \\
\midrule
Random & 1.13 & 3.17 & 0.0\% \\
Flat Checklist & 1.76 & 2.43 & 23.3\% \\
CDG Topological Order & 3.67 & 1.58 & 50.2\% \\
CDG + Severity Weighting & \textbf{4.16} & \textbf{1.38} & \textbf{56.5\%} \\
\bottomrule
\end{tabular}
\caption{Upstream-Priority Fix Efficiency (RCFE, measured via HyTPS-Bench). CDG-aware repair targets upstream violations first, reducing iterations and downstream rework.}
\label{tab:cdg_rcfe}
\end{table}

\subsection{Prior-Constrained Constraint Dependency Graph}
Traditional engineering constraint checking relies on flat rule lists. The proposed framework upgrades this to a \textbf{prior-constrained, empirically calibrated constraint dependency graph} \(G=(V,E)\). Crucially, the graph structure is not discovered via causal induction; rather, edge \emph{directions} are constrained by a domain-informed lifecycle ordering prior (unit \(\to\) physical \(\to\) numerical \(\to\) execution \(\to\) audit). This encodes the fundamental engineering principle that upstream constraints must be satisfied before downstream ones become physically meaningful.

Within this fixed structure, edge \emph{weights} are calibrated from repair episodes in historical VER loops. Edge weights \(w_{u\to v}(c)\) represent the \textbf{empirical conditional probability} that resolving upstream node \(u\) also resolves downstream node \(v\), given the current regime context \(c\). These weights provide a data-informed refinement of the lifecycle prior, allowing the Agent to prioritize root-cause nodes with the largest estimated downstream repair gain (Eq.~\ref{eq:cdg_priority}). It is important to emphasize that these weights represent association-based statistics from repair traces, not causal relationships.

Edge weights \(w_{u\to v}(c)\) are \textbf{context-conditioned} on the current flight/thermal regime \(c\) 
(e.g., extreme heat flux vs.\ nominal re-entry), estimated as the empirical conditional probability that resolving upstream node \(u\) also resolves downstream node \(v\):
\[
w_{u\to v}(c) = P(v_{t+1}\!=\!1 \mid v_t\!=\!0,\; u\text{ repaired},\; c).
\]
At deployment, the agent prioritizes root-cause nodes with the largest estimated downstream repair gain:
\begin{equation}\label{eq:cdg_priority}
u^* = \arg\max_{u\in\text{Violations}} \sum_{v\in\text{Descendants}(u)} w_{u\to v}(c).
\end{equation}
This supports upstream-priority repair based on empirical dependency structure rather than heuristic symptom patching. In this paper, ``root cause'' is an operational notion denoting upstream fault candidates in the CDG topological order, not a claim of causal identification.

\paragraph{CDG construction.}
The CDG is calibrated from \textbf{4,206 repair episodes} collected across 483 simulation tasks during VER loop execution. Each episode records a structured tuple $(s_t, a_t^{\text{repair}}, s_{t+1}, \Delta V_t)$, where $\Delta V_t$ captures changes in the violation bitmap after targeting a specific constraint node $u$. When fixing $u$ also resolves downstream node $v$, this provides evidence for a directed edge $u\to v$. Crucially, edge orientations are \emph{constrained by a lifecycle ordering prior} (unit $\to$ physical $\to$ numerical $\to$ execution $\to$ audit) that encodes engineering domain knowledge; learned edge weights refine this prior-constrained structure using empirical co-resolution frequencies, rather than discovering dependencies from scratch. Accordingly, CDG edge weights represent \emph{association-based statistics} from repair traces---empirical conditional probabilities of co-resolution---and should not be interpreted as causal relationships. Edge weights are estimated as empirical conditional propagation probabilities stratified by four regime contexts (nominal re-entry, moderate ablation, high heat flux, extreme thermal-chemical). The resulting CDG has \textbf{23 nodes} and \textbf{45 directed edges}; validation on a 20\% held-out split ($n=841$) yields a mean absolute calibration error of 0.08 in edge weight predictions.

\paragraph{Motivation for CDG repair ordering.}
The priority score (Eq.~\eqref{eq:cdg_priority}) is the greedy one-step maximizer of expected downstream violations resolved. In our CDG, unit and physics nodes account for 71\% of total downstream repair gain (propagation ratio $\hat{\rho}\approx 2.9$ on held-out data), meaning CDG ordering resolves substantially more violations per repair action than flat checklist---consistent with the observed RCFE ratio (4.16 CDG vs.\ 1.76 flat checklist, Table~\ref{tab:cdg_rcfe}). Severity weighting (Table~\ref{tab:cdg_rcfe}, last row) further extends the objective with severity-weighted downstream gain. These observations are engineering motivation rather than formal approximation guarantees. Importantly, a substantial share of this structure is inherited from the lifecycle ordering prior; the learned edge weights provide a data-informed refinement of that prior rather than causal discovery. The main limitation is miscalibration in deep multi-layer cascades (5.2\% failure rate, Section~\ref{sec:results}).

\subsection{AeroTherm-GPT: TPS-Oriented Instantiation of CCLG}
AeroTherm-GPT is a TPS-specialized system built on the CCLG framework. Its implementation combines domain adaptation, retrieval, verifier-guided repair, and preference alignment, as shown in Fig.~\ref{fig:c2lg_architecture}.

\subsubsection{Outcome-Supervised Engineering Reasoning Traces}
In the SFT stage, AeroTherm-GPT learns to generate not only final code
but also structured \textbf{Engineering Reasoning Traces} (Design Memos),
a domain-specific chain-of-thought \cite{wei2022chain,kojima2022large} that records problem decomposition, modeling choices, and regime conditions.

We use \textbf{outcome-supervised data filtering}: 
only trajectories that pass all VER gates are used for training. 
The loss is:
\[
\mathcal{L}_{\text{OS-SFT}} = -\sum_{i=1}^N \mathbb{1}\{\text{Ready}(y_i)=1\} \cdot \log P_\theta(y_i\mid x_i).
\]
This encourages the model to learn physically valid, executable reasoning rather than superficial expert mimicry.

\subsubsection{Constraint-Guided Search via VER Loop}
The VER loop implements a \textbf{constraint-guided tree search} with test-time compute scaling \cite{snell2025scaling,brown2024large}. A lightweight \textbf{Process Reward Model (PRM)} \cite{uesato2022solving,lightman2024verify} \(V_\theta(s_t) \approx \mathbb{E}_\pi[\mathbb{1}\{\text{Ready}(s_T){=}1\} \mid s_t]\) scores intermediate repair states. At each step, the leaf with highest PRM score is expanded: CDG-ordered violations are extracted, violation-triggered RAG retrieves repair guidance, and $K{=}4$ candidate repairs are sampled and scored. The search terminates when $\text{Ready}(s_t){=}1$ or the compute budget (8 iterations) is exhausted (full algorithm pseudocode available at \url{https://github.com/TPS-qxx/AeroTherm-GPT}).

\subsubsection{Constraint-Decomposed Preference Optimization}
We perform \textbf{constraint-decomposed DPO} \cite{rafailov2023dpo,ouyang2022training} using three levels of automatically constructed preferences:
\begin{enumerate}
    \item \textbf{Binary pass signal}: \(\text{Ready}(y_w)=1 \succ \text{Ready}(y_l)=0\)
    \item \textbf{Self-evolution}: post-repair \(y_w \succ\) pre-repair \(y_l\)
    \item \textbf{Severity-graded preference}: \(R(y_w) > R(y_l) \Rightarrow y_w \succ y_l\)
\end{enumerate}

The DPO objective is:
\[
\mathcal{L}_{\text{DPO}} = -\mathbb{E}_{(x,y_w,y_l)}\left[
\log\sigma\!\left(
\beta\log\frac{\pi_\theta(y_w\mid x)}{\pi_{\text{ref}}(y_w\mid x)}
-\beta\log\frac{\pi_\theta(y_l\mid x)}{\pi_{\text{ref}}(y_l\mid x)}
\right)
\right].
\]
The model learns to avoid high-severity cascading physical violations, improving first-pass success.

\section{Evaluation Protocol}\label{sec:benchmark}
We evaluate the proposed framework along two parallel and complementary tracks. \textbf{Track~A (Domain-Specific)}: hypersonic domain evaluation comprising two complementary sub-tracks---\textbf{Track~A-1 (HyTPS-Bench)}, a workflow-aligned benchmark grounded in real TPS engineering practice that tests executable simulation artifact generation under multi-gate engineering constraints, and \textbf{Track~A-2 (HyperKFA-Bench~\cite{zheng2025hyperkfa})}, the first benchmark covering hypersonic fundamental knowledge, formula invocation, and automated programming, providing evidence within the hypersonic domain beyond our workflow-specific setting. \textbf{Track~B (Cross-Benchmark)}: a suite of external benchmarks tests whether the framework's core mechanisms---domain-adapted reasoning, iterative repair---transfer to general scientific and programming domains.

The two tracks serve complementary evidential roles. Track~A measures hypersonic domain capability across two dimensions: engineering workflow utility (Track~A-1, high ecological validity) and fundamental scientific knowledge and programming competence (Track~A-2, beyond our workflow-specific setting). Track~B provides broader generalization evidence on scientific reasoning and code generation at the cost of lower domain specificity. Together, the tracks establish that the framework's gains are genuine, domain-grounded, and not confined to the self-developed evaluation setting.

\subsection{Track A-1: HyTPS-Bench --- Workflow-Aligned Domain Evaluation}
\label{subsec:hytpsbench}

HyTPS-Bench is a domain benchmark grounded in real hypersonic TPS engineering practice. Its task definitions, problem templates, and ground-truth answers are derived from publicly available engineering literature, de-identified arc-jet test records, and physics-based simulation oracles---none of which originate from or overlap with AeroTherm-GPT's training data. The benchmark is not constructed to favor any particular system architecture: task specifications are written from engineering requirements (``given this geometry, material, and boundary condition, produce an executable simulation''), and correctness is judged by a sandboxed Reference Executor and physics oracle that operate externally to the CCLG code generator. This construction ensures that HyTPS-Bench measures genuine TPS engineering capability rather than familiarity with a specific system's output format or training distribution.

The benchmark design is motivated by a coverage gap in existing evaluation resources: general benchmarks (MMLU, SciBench, HumanEval) measure either knowledge breadth or code generation in isolation, but none covers the full Problem$\to$Physical$\to$Numerical$\to$Implementation$\to$Verification workflow with unit-sensitive and functionally correct evaluation that industrial TPS practice demands. HyTPS-Bench addresses this gap by integrating all five workflow stages with sandboxed execution and audit traceability.

\subsubsection{Tasks and Engineering Metrics}

HyTPS-Bench covers four tasks that collectively mirror the TPS engineering workflow. Tasks T1 and T2 test knowledge-grounding capabilities; T3 and T4 test executable and auditable artifact generation. The primary evaluation metrics are:
\begin{itemize}
    \item \textbf{EESR} (End-to-End Success Rate): binary gate requiring a code package to pass all verification, execution, and audit checks simultaneously.
    \item \textbf{RCFE} (Root-Cause Fix Efficiency): downstream violations resolved per root-cause repair action.
    \item \textbf{ACS} (Audit Completeness Score): traceability of the generated artifacts.
\end{itemize}

\noindent\textbf{Task design and motivation.}
\textbf{T1 (Material Property QA)} targets a foundational engineering operation: retrieving material properties under explicitly stated conditions (material, regime, target property). It probes whether a model can provide precise, context-dependent facts while maintaining unit fidelity---for instance, correctly returning the maximum service temperature of a C/C composite in an oxidizing atmosphere as 1650\,°C rather than a plausible but incorrect value.
\textbf{T2 (Technical Document Analysis)} evaluates long-context comprehension and structured extraction from dense technical passages (0.5--2 pages, 500--1500 tokens), including a cross-lingual EN/CN subset for multilingual stress testing.
\textbf{T3 (Simulation Input Generation)} measures the pivotal capability to translate a natural-language physical scenario into an executable Python script using standard scientific libraries. Prompts specify slab geometry, material properties, and boundary/initial conditions; correct solutions compute temperature evolution and report values at designated points and times under 1D transient heat conduction, with sandboxed execution and comparison against a physics oracle.
\textbf{T4 (Post-Processing + Lifecycle Audit)} assesses downstream artifact completeness: executing analysis scripts, producing required output artifacts, and maintaining a traceable provenance record.

\noindent\textbf{Task-specific scoring.}
T1 uses unit-aware Exact Match (EM $\pm 5\%$ tolerance $\times$ UC), where UC checks unit string equivalence (e.g., \texttt{degC} matches \texttt{Celsius}). T2 reports micro-F1 on structured parameter extraction and ROUGE-L/BERTScore on summarization. T3 defines two complementary scores: \emph{SI-G} $=$ Spec-Comp $\times$ Ref-Exec $\times$ Func, a backend-agnostic functional score where $y^\star$ is produced by a deterministic reference executor (conduction oracle, pass@1 $\pm 2\%$ tolerance), and \emph{EESR} $=$ $\mathbb{1}$[backend executes] $\times$ $\mathbb{1}$[validators pass] $\times$ $\mathbb{1}$[audit complete], requiring all five gate categories to pass simultaneously. T4 uses PP-A $=$ Exec $\times$ Artifacts $\times$ ACS. Scoring pseudo-code and the sandboxed execution environment are provided in the released evaluation scripts.

\subsubsection{Dataset Construction and Governance}\label{subsec:data_governance}
HyTPS-Bench contains approximately 2,500 JSON instances ($T_1$/$T_2$/$T_3$/$T_4 = 25\%/25\%/30\%/20\%$; Table~\ref{tab:dataset_composition_distribution}). Instances are drawn from three sources: (1)~public literature (NASA NTRS, AIAA journals) covering TPS materials, test procedures, and analysis techniques; (2)~de-identified experimental snippets (arc-jet test conditions, thermocouple records) with all sensitive or export-controlled content scrubbed; and (3)~physics-based synthetic problems generated by the Reference Executor with known ground-truth fields. The cross-lingual subset ($q \approx 0.2$) uses parallel EN/CN technical texts. Publicly released benchmark materials and utilities are available in the project repository (\url{https://github.com/TPS-qxx/AeroTherm-GPT}).

\subsubsection{Benchmark Difficulty and Baseline Performance}
To establish that HyTPS-Bench is genuinely challenging and not saturated by existing models, we evaluate ten state-of-the-art LLMs in zero-shot setting prior to AeroTherm-GPT evaluation. Results confirm a systematic bottleneck: while models achieve strong factual recall on T1 (best MP-QA: 1.00 by DeepSeek-V3/R1) and moderate document comprehension on T2, simulation code generation (T3) is universally difficult---the best zero-shot SI-G pass@1 is 0.55 (Claude-4.1-Opus), and several capable models including Qwen3-32B and DeepSeek-V3 score poorly (below 0.40). This confirms that the SI-G task, which requires multi-step physics reasoning, unit consistency, and boundary condition handling simultaneously, exposes a critical gap between declarative knowledge and functionally correct engineering code generation that current LLMs have not closed without domain-specific adaptation.

\subsubsection{Evaluation Integrity Controls}\label{subsec:leakage}
Because CCLG and HyTPS-Bench were developed within the same project, we enforce isolation at the level of \emph{instances, templates, calibration episodes, and scoring} (as detailed in Table~\ref{tab:leakage_control}). Specifically, benchmark test instances are never used as SFT traces, DPO preference pairs, PRM rollouts, or CDG calibration data; CDG repair episodes are collected from a separate task pool and are excluded from benchmark test construction; synthetic benchmark instances are generated from held-out parameter ranges/material splits relative to CDG calibration; and scoring is performed by an externally implemented Reference Executor and validator stack that are not part of the AeroTherm-GPT generator. The only shared component across systems is the \emph{constraint asset schema/API}: all baselines can access the same constraint definitions, but not privileged repair trajectories, hidden labels, or benchmark-specific oracle outputs. We acknowledge that EESR's multiplicative gate structure structurally favors systems with explicit verification scaffolds, which is ecologically valid for safety-critical deployment; T1/T2 knowledge-track scores and Track~B cross-benchmarks provide signals beyond the workflow-specific setting.

\begin{table}[!t]
\centering
\small
\setlength{\tabcolsep}{4pt}
\renewcommand{\arraystretch}{1.2}
\begin{tabularx}{\textwidth}{@{}l c c c X@{}}
\toprule
\textbf{Component} & \textbf{Training} & \textbf{Evaluation} & \textbf{Shared} & \textbf{Leakage control} \\
\midrule
Constraint assets (schema) & Yes & Yes & Yes (API) & Shared schema/API only; no privileged repair traces or oracle outputs \\
Reference Executor & No & Yes & Yes & External scoring implementation, identical for all systems \\
Repair logs (CDG training) & Yes & No & No & Separate calibration pool; excluded from benchmark test construction \\
HyTPS-Bench test set & No & Yes & Yes & Held out at benchmark-instance and template-family level \\
Validator rules & No & Yes (post) & Yes & Applied uniformly post-generation to every system \\
Synthetic instance parameters & Subset & Remaining & N/A & Held-out parameter ranges with material-class-disjoint split \\
\bottomrule
\end{tabularx}
\caption{Data leakage and evaluation integrity control. Each component's role in training vs.\ evaluation is explicitly documented.}
\label{tab:leakage_control}
\end{table}

\begin{table}[!t]
\centering
\small
\renewcommand{\arraystretch}{1.1}
\setlength{\tabcolsep}{4pt}
\begin{tabular}{@{}p{0.35\linewidth}p{0.55\linewidth}@{}}
\toprule
Item & HyTPS-Bench Value \\
\midrule
Size & $\sim$2,500 JSON instances \\
Task mixture (T1--T4) & 25\%/25\%/30\%/20\% \\
Sources & Public lit.\ (NASA NTRS, AIAA); de-identified snippets; synthetic \\
Multilingual & EN/CN, $q \approx 0.2$ \\
Input length & 0.5--2 pages (500--1500 tokens) \\
Token stats & $\mu_{\text{tok}} \approx 900$, $p95_{\text{tok}} \approx 1500$ \\
MP-QA tolerance & EM $\pm 5\%$ + unit equivalence \\
SI-G tolerance & pass@1 $\pm 2\%$ of oracle \\
Best zero-shot SI-G (10 baselines) & 0.55 (Claude-4.1-Opus); Qwen3-32B / DeepSeek-V3: < 0.40 \\
License & CC BY-NC-SA 4.0 \\
\bottomrule
\end{tabular}
\caption{Dataset composition and baseline difficulty facts for HyTPS-Bench. The zero-shot SI-G results confirm the benchmark is not saturated by existing models.}
\label{tab:dataset_composition_distribution}
\end{table}

\subsection{Track A-2: HyperKFA-Bench --- Hypersonic Knowledge and Programming Evaluation}
To provide evidence of hypersonic domain competence beyond the workflow-specific setting, we evaluate on \textbf{HyperKFA-Bench} (\textbf{Hyper}sonic \textbf{K}nowledge, \textbf{F}ormula invocation, and \textbf{A}utomated programming Benchmark)~\cite{zheng2025hyperkfa}. HyperKFA-Bench is the first benchmark designed to assess hypersonic and TPS domain competence. It serves as a crucial validation of the domain-specific knowledge acquired by AeroTherm-GPT, distinct from the workflow-oriented evaluation of HyTPS-Bench.

\begin{enumerate}
    \item \textbf{Fundamental Knowledge} (30 questions, Q1--Q30): factual recall of core hypersonic aerothermodynamics, gas dynamics, and TPS-relevant concepts. Questions span Mach number thresholds, shock wave behavior, boundary layer heat transfer, catalytic wall effects, ablation failure mechanisms, rarefied gas phenomena, and thermal protection material properties.
    \item \textbf{Formula Invocation} (6 tasks, F1--F6): quantitative calculation tasks requiring correct identification and application of governing equations under specified flight conditions, including oblique shock angle ($\theta$-$\beta$-Mach relation), stagnation point heat flux (Sutton--Graves formula), hypersonic Reynolds number, Prandtl--Meyer expansion angle, ablation layer thickness, and radiative equilibrium temperature.
    \item \textbf{Automated Programming} (4 tasks, P1--P4): Python code generation tasks of increasing complexity, covering 1D Riemann problem solving (Euler equations), catalytic recombination coefficient fitting via machine learning, 2D plain-weave textile structure generation, and high-temperature chemical equilibrium species calculation for a five-component air model.
\end{enumerate}

\noindent The benchmark thus probes knowledge depth, formula fluency, and scientific code generation in a standalone, zero-shot setting---capabilities that underpin but are distinct from the full engineering artifact generation workflow measured by HyTPS-Bench.

\paragraph{Evaluation protocol.}
For Fundamental Knowledge, correctness is determined by exact-match comparison against ground-truth answers. For Formula Invocation, a response is marked correct if the numerical result falls within 2\% relative error of the reference value. For Automated Programming, generated code is compiled and executed in Python~3.9; a task is successful if the code runs without error and produces outputs consistent with the reference solution. The overall \textbf{Task Completion Rate} (TCR) is the fraction of all 40 tasks completed correctly. Baseline results for DeepSeek-R1, DeepSeek-V3, GPT-4o, and Qwen2.5 are taken directly from the original publication~\cite{zheng2025hyperkfa}. AeroTherm-GPT is evaluated on the same prompts under zero-shot conditions without TPS-specific scaffolding (no VER loop, no RAG retrieval), so that results reflect the competence acquired through domain SFT and DPO rather than the deployment-time repair infrastructure.

\subsection{Track B: Cross-Benchmark Validation Protocol}
\label{subsec:crossbench}

To provide evidence that the framework's core mechanisms generalize beyond HyTPS-Bench, we evaluate on five external benchmarks spanning scientific reasoning and code generation.

\paragraph{Scientific reasoning and knowledge benchmarks.}
These benchmarks test whether domain-adapted SFT preserves and enhances general scientific reasoning, and whether it causes catastrophic forgetting on broader capabilities:
\begin{itemize}
\item \textbf{SciBench} \cite{wang2024scibench}: 695 college-level open-ended science problems (physics, chemistry, mathematics). Metric: answer accuracy.
\item \textbf{GPQA-Diamond} \cite{rein2024gpqa}: 198 graduate-level STEM questions designed to be resistant to web search. Metric: accuracy.
\item \textbf{MMLU-STEM} \cite{hendrycks2021measuring}: 3,014 multiple-choice questions across 18 STEM subjects. Metric: 5-shot accuracy.
\end{itemize}

\paragraph{Code generation benchmarks with iterative repair.}
These benchmarks test whether CCLG's iterative repair architecture---stripped of TPS-specific constraint definitions---improves general code generation. On these tasks, the VER loop operates with execution-based feedback (run code $\to$ check test output $\to$ provide error traceback $\to$ repair), using a flat repair strategy without CDG ordering (since general code tasks lack the engineering constraint hierarchy that CDG models):
\begin{itemize}
\item \textbf{HumanEval} \cite{chen2021evaluating}: 164 Python function synthesis tasks with unit tests. Metric: pass@1.
\item \textbf{MBPP} \cite{austin2021program}: 500 entry-level Python programming tasks. Metric: pass@1.
\end{itemize}

\noindent For code benchmarks, we evaluate each model in two configurations: (1)~\emph{1-pass}: single-shot generation without repair; and (2)~\emph{+VER}: up to 3 iterative repair rounds with execution feedback, using the same VER loop infrastructure as CCLG but without domain-specific CDG ordering. This isolates the contribution of the \emph{iterative repair architecture} from domain-specific constraint prioritization.

\subsection{Baselines and Implementation Details}\label{subsec:baselines}

\subsubsection{Implementation Details}
AeroTherm-GPT follows a three-stage training pipeline: (1)~outcome-supervised SFT on 8,500 domain-specific engineering reasoning traces using LoRA \cite{hu2022lora,touvron2023llama}; (2)~constraint-decomposed DPO \cite{rafailov2023dpo} on 4,206 repair episodes; and (3)~PRM training on 12,400 (state, outcome) pairs from VER-loop rollouts. At inference, the VER tree search uses beam width $K{=}4$ and a maximum budget of 8 repair iterations with best-first PRM selection. All training and inference were conducted on a cluster of 8 NVIDIA A100 (80GB) GPUs. Implementation details sufficient to run the released deployment pipeline, benchmark subset utilities, tests, sample data, and demo are provided in the open-source repository (\url{https://github.com/TPS-qxx/AeroTherm-GPT}). Trained weights, the full training pipeline, and the internal retrieval/reward-model stack are not publicly released.

\subsubsection{Baselines}
We evaluate closed-source (GPT-5, Claude-4, GPT-4o, Claude-3.5) and open-source (Qwen3, DeepSeek-V3/R1, GLM-4.5) models under three prompting levels: \textbf{Zero-Shot}, \textbf{Reasoning} (test-time compute scaling), and \textbf{RAG + Few-Shot}. These establish a no-workflow-adaptation lower bound; the structurally informative comparisons are among scaffolded repair systems (Rows 4--5 in Table~\ref{tab:main_results}) sharing identical tool access. Four scaffolded repair baselines isolate specific contributions: \textbf{GPT-4o+Refl.} (symptom-based repair), \textbf{Tool-Agent} (same tools, no CDG/SFT/DPO), \textbf{Flat-VER} (same VER loop with flat checklist ordering), and \textbf{CDG-SFT Agent} (CDG ordering + SFT, no DPO). Table~\ref{tab:baseline_capabilities} summarizes per-baseline capabilities.

\begin{table}[!t]
\centering
\scriptsize
\renewcommand{\arraystretch}{1.0}
\setlength{\tabcolsep}{3pt}
\begin{tabular}{@{}lcccccc@{}}
\toprule
\textbf{Method} & \textbf{Retrieval} & \textbf{Validator feedback} & \textbf{Exec. backend} & \textbf{Iterative repair} & \textbf{CDG ordering} & \textbf{Domain SFT+DPO} \\
\midrule
Zero-Shot LLMs        & \xmark & \xmark & \xmark & \xmark & \xmark & \xmark \\
Reasoning (w/ think)  & \xmark & \xmark & \xmark & \xmark & \xmark & \xmark \\
RAG + Few-Shot        & \cmark & \xmark & \xmark & \xmark & \xmark & \xmark \\
GPT-4o+Refl.          & \xmark & \cmark & \cmark & \cmark & \xmark & \xmark \\
Tool-Agent            & \cmark & \cmark & \cmark & \cmark & \xmark & \xmark \\
Flat-VER              & \cmark & \cmark & \cmark & \cmark & \xmark & \xmark \\
Flat-VER + SFT + DPO  & \cmark & \cmark & \cmark & \cmark & \xmark & \cmark \\
CDG-SFT Agent         & \cmark & \cmark & \cmark & \cmark & \cmark & SFT only \\
\textbf{AeroTherm-GPT (Full)} & \cmark & \cmark & \cmark & \cmark & \cmark & \cmark \\
\bottomrule
\end{tabular}
\caption{Baseline capability alignment. ``Flat-VER + SFT + DPO'' and ``CDG-SFT Agent'' are symmetric ablation baselines: the former adds SFT+DPO to flat repair (isolating CDG's contribution); the latter adds CDG ordering to SFT-only training (isolating DPO's contribution). AeroTherm-GPT combines both.}
\label{tab:baseline_capabilities}
\end{table}

Full prompt templates are provided in the released source code.

\section{Results}\label{sec:results}
We organize results along the two evaluation tracks defined in Section~\ref{sec:benchmark}. Section~\ref{subsec:main_results} reports Track~A-1 (HyTPS-Bench) results; Section~\ref{subsec:pofbench_results} reports Track~A-2 (HyperKFA-Bench) results; Section~\ref{subsec:crossbench_results} reports Track~B (cross-benchmark validation). Subsequent subsections present ablation, backbone transfer, and failure analysis.

\subsection{Track A-1: Main Results on HyTPS-Bench}\label{subsec:main_results}

\begin{figure}[!t]
    \centering
    \includegraphics[width=0.95\linewidth]{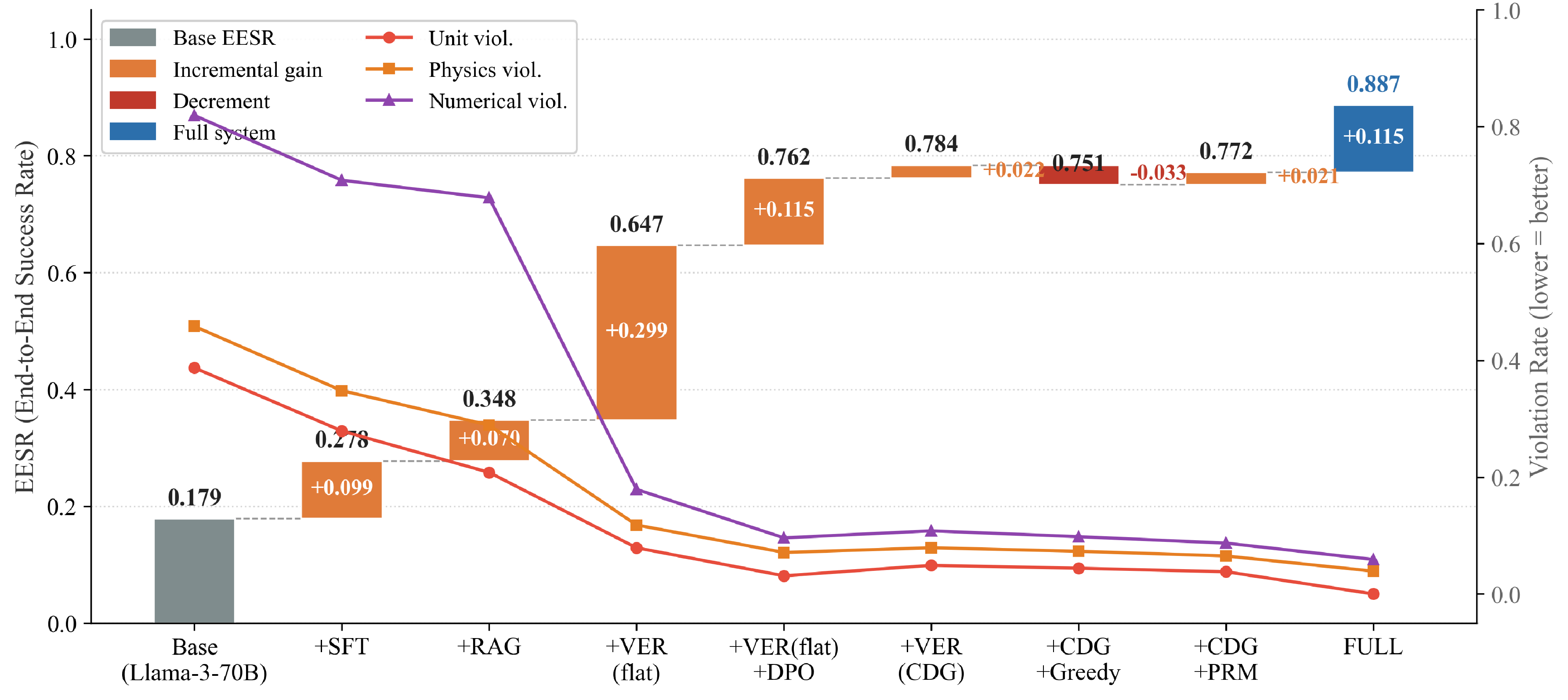}
    \caption{Ablation Study --- Incremental EESR Contribution per Module with Violation Rate Trajectories. The waterfall chart decomposes the EESR gain from each framework component, while the overlaid lines track the corresponding reduction in unit, physics, and numerical violation rates.}
    \label{fig:ablation_study}
\end{figure}

Table~\ref{tab:main_results} summarizes results (all CIs: bootstrap $B{=}1000$; pairwise tests: McNemar with Holm--Bonferroni correction). AeroTherm-GPT achieves \textbf{88.7\% EESR} [87.5--89.9], compared with 48.0\% for GPT-5 High, 78.4\% for CDG-SFT Agent, and 46.8\% for Flat-VER. CDG-SFT Agent (CDG + SFT, no DPO) reaches 78.4\% EESR, while Flat-VER + SFT + DPO (flat repair + SFT + DPO, no CDG) reaches 76.2\% EESR, confirming that CDG ordering and constraint-decomposed DPO each contribute to performance. Comparing either ablation row with AeroTherm-GPT isolates the marginal gain from adding the missing component: $+10.3$~pp from DPO (CDG-SFT Agent $\to$ AeroTherm-GPT, $p<0.001$) and $+12.5$~pp from CDG ordering (Flat-VER + SFT + DPO $\to$ AeroTherm-GPT, $p<0.001$). The component-level ablation (Fig.~\ref{fig:ablation_study}) further shows positive contributions from SFT, CDG ordering, PRM-guided search, and DPO. Along the matched incremental path shown in the figure, CDG ordering contributes $+9.1$~pp ($p<0.01$), PRM-guided search $+3.4$~pp ($p=0.03$), and DPO $+11.5$~pp ($p<0.001$); the exact incremental gain attributed to SFT is $+17.9$~pp, successfully closing the gap between the un-finetuned Flat-VER baseline (46.8\%) and the sum of the scaffold components (24.0~pp). These differences between the incremental ablation steps and the baseline-comparison figures reflect interaction effects between simultaneously active components. CDG and DPO are complementary: CDG defines repair structure at deployment; DPO internalizes dependency-aware preferences into first-pass generation.

\begin{table}[!t]
\centering
\small
\renewcommand{\arraystretch}{1.0}
\setlength{\tabcolsep}{3pt}

\begin{tabular*}{\textwidth}{@{\extracolsep{\fill}}lcccccccc@{}}
\toprule
\textbf{Model} &
\multicolumn{2}{c}{\textbf{Core}} &
\multicolumn{3}{c}{\textbf{Task Perf.}} &
\textbf{Over.} &
\multicolumn{2}{c}{\textbf{Cons. (\%)}} \\
\cmidrule(lr){2-3} \cmidrule(lr){4-6} \cmidrule(lr){8-9}
& \textbf{EESR} & \textbf{SI-G} & \textbf{MP} & \textbf{TD} & \textbf{PP-A} & & \textbf{U/P} & \textbf{N} \\
\midrule
\rowcolor{gray!15} \multicolumn{9}{l}{\textit{1. Baseline LLMs (w/o thinking)}} \\
GPT-5 (min)            & 0.350 & 0.290 & 0.641 & 0.573 & 0.521 & 0.475 & 62.3 & 28.5 \\
Claude-4-Sonnet (w/o)  & 0.420 & 0.480 & 0.493 & 0.712 & 0.643 & 0.550 & 68.2 & 32.1 \\
Claude-4.1-Opus (w/o)  & 0.450 & 0.550 & 0.551 & 0.741 & 0.691 & 0.597 & 70.5 & 33.7 \\
Qwen3-32B (w/o)        & 0.250 & 0.230 & 0.387 & 0.331 & 0.293 & 0.298 & 51.7 & 22.4 \\
DeepSeek-V3-0324       & 0.320 & 0.360 & 0.571 & 0.504 & 0.452 & 0.441 & 58.9 & 26.8 \\
GLM-4.5 (w/o)          & 0.380 & 0.500 & 0.452 & 0.397 & 0.373 & 0.420 & 60.3 & 29.2 \\
\midrule
\rowcolor{gray!15} \multicolumn{9}{l}{\textit{2. Baseline LLMs (w/ thinking)}} \\
GPT-5 (high)           & 0.480 & 0.540 & 0.672 & 0.618 & 0.568 & 0.576 & 72.8 & 35.9 \\
Claude-4-Sonnet (w/)   & 0.450 & 0.530 & 0.768 & 0.421 & 0.401 & 0.514 & 65.7 & 31.2 \\
Claude-4.1-Opus (w/)   & 0.470 & 0.540 & 0.803 & 0.487 & 0.457 & 0.551 & 68.3 & 33.1 \\
Qwen3-235B-T           & 0.400 & 0.480 & 0.562 & 0.497 & 0.463 & 0.480 & 63.2 & 28.7 \\
DeepSeek-R1-0528       & 0.370 & 0.430 & 0.381 & 0.317 & 0.308 & 0.361 & 59.1 & 26.4 \\
GLM-4.5 (w/)           & 0.420 & 0.540 & 0.421 & 0.364 & 0.341 & 0.417 & 61.5 & 29.8 \\
\midrule
\rowcolor{gray!15} \multicolumn{9}{l}{\textit{3. Enhanced Baselines}} \\
GPT-4o (Zero Shot)     & 0.281 & 0.427 & 0.812 & 0.753 & 0.671 & 0.589 & 67.8 & 31.7 \\
C3.5 (Zero Shot)       & 0.307 & 0.483 & 0.843 & 0.779 & 0.706 & 0.624 & 70.9 & 34.8 \\
GPT-4o (RAG+Few Shot)  & 0.452 & 0.591 & 0.871 & 0.824 & 0.754 & 0.698 & 80.7 & 37.9 \\
\midrule
\rowcolor{gray!15} \multicolumn{9}{l}{\textit{4. Scaffolded Repair Baselines (same tool access)}} \\
GPT-4o+Refl.           & 0.376 & 0.537 & 0.831 & 0.774 & 0.751 & 0.729 & 75.8 & 43.7 \\
Tool-Agent             & 0.441 & 0.612 & 0.847 & 0.788 & 0.762 & 0.743 & 78.3 & 46.1 \\
Flat-VER               & 0.468 & 0.638 & 0.854 & 0.803 & 0.774 & 0.761 & 81.2 & 48.9 \\
Flat-VER + SFT + DPO   & 0.762 & 0.731 & 0.891 & 0.843 & 0.813 & 0.819 & 88.6 & 63.7 \\
CDG-SFT Agent          & 0.784 & 0.714 & 0.883 & 0.836 & 0.804 & 0.804 & 87.4 & 61.3 \\
\midrule
\rowcolor{gray!15} \multicolumn{9}{l}{\textit{5. Our Proposed System}} \\
\textbf{AeroTherm-GPT} & \textbf{0.887} & \textbf{0.847} & \textbf{0.924} & \textbf{0.871} & \textbf{0.843} & \textbf{0.879} & \textbf{96.8} & \textbf{93.7} \\
\bottomrule
\end{tabular*}
\caption{Main results on HyTPS-Bench. Core: EESR = End-to-End Success Rate, SI-G(T3) = Simulation Input Generation. Task Perf.: MP = Material Property QA (T1), TD = Technical Document Analysis (T2), PP-A = Post-Processing + Lifecycle Audit (T4). Over. = overall; U/P = unit/physics compliance; N = numerical compliance (\%).}
\label{tab:main_results}
\end{table}

\begin{figure}[!ht]
    \centering
    \includegraphics[width=0.8\linewidth]{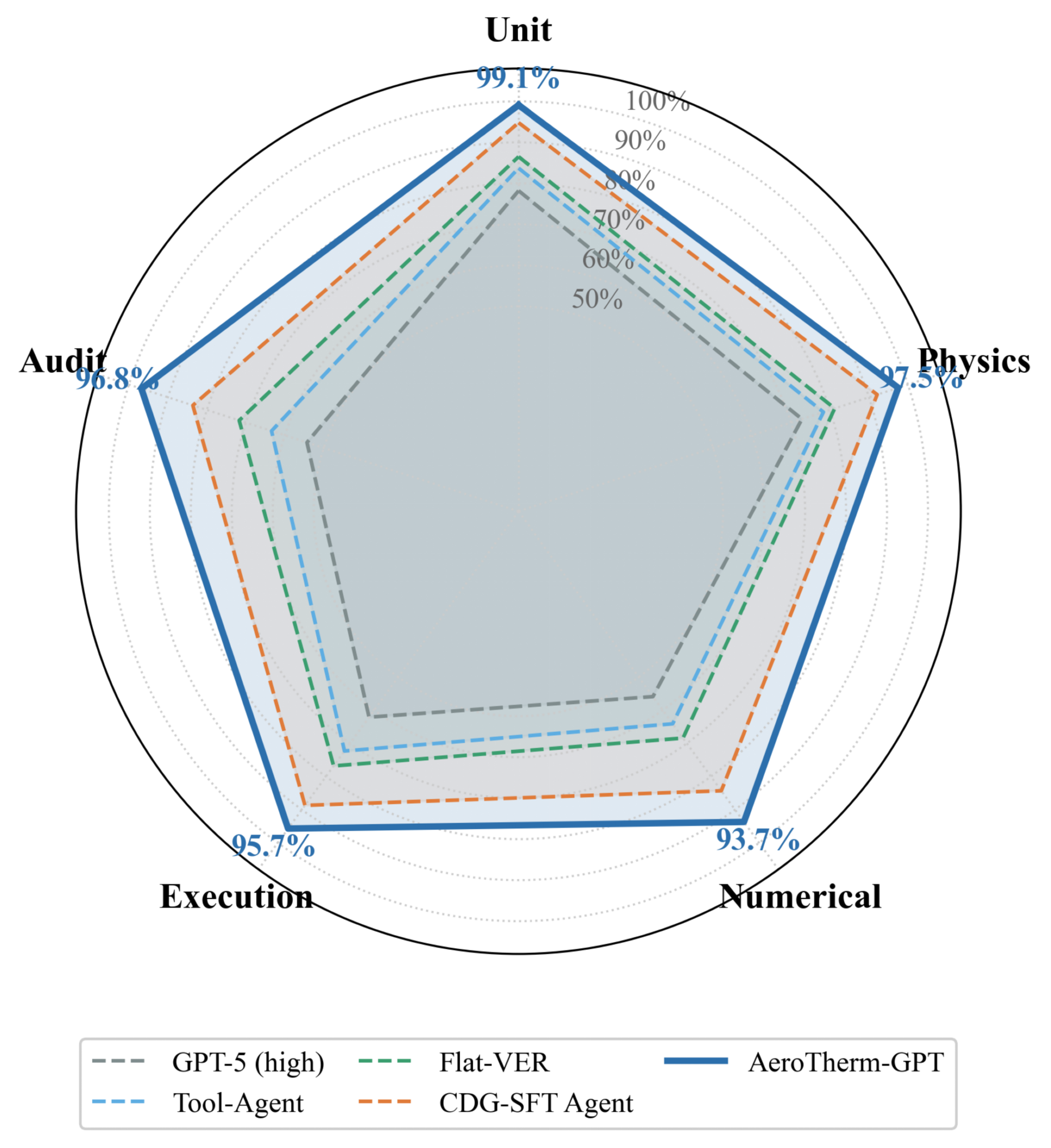}
    \caption{Per-Gate Pass Rates --- Five Methods $\times$ Five Verification Gates. The radar chart visualizes the compliance rate across unit, physics, numerical, execution, and audit gates, highlighting AeroTherm-GPT's near-ceiling performance across all categories.}
    \label{fig:per_gate_radar}
\end{figure}

\paragraph{Per-gate decomposition.}
Figure~\ref{fig:per_gate_radar} reports individual gate pass rates. AeroTherm-GPT achieves near-ceiling unit (99.1\%) and physics (97.5\%) compliance; DPO's largest gains over CDG-SFT Agent are in numerical (+9.4~pp) and audit (+13.2~pp) gates.

\paragraph{Computational cost.}
AeroTherm-GPT's multi-step VER loop incurs higher inference cost than single-pass baselines. On HyTPS-Bench T3 tasks, the average wall-clock time per task is 4.2 minutes (including VER iterations), compared with 0.8 minutes for zero-shot GPT-4o and 2.6 minutes for Flat-VER. The average number of LLM calls per task is 3.8 (including the initial generation and 0.58 average repair iterations after DPO alignment). Token consumption averages 12,400 input + 8,600 output tokens per task. This overhead is acceptable for offline engineering design (where manual workflows require hours), but may require optimization for time-sensitive applications.

\begin{figure}[!ht]
    \centering
    \includegraphics[width=0.95\linewidth]{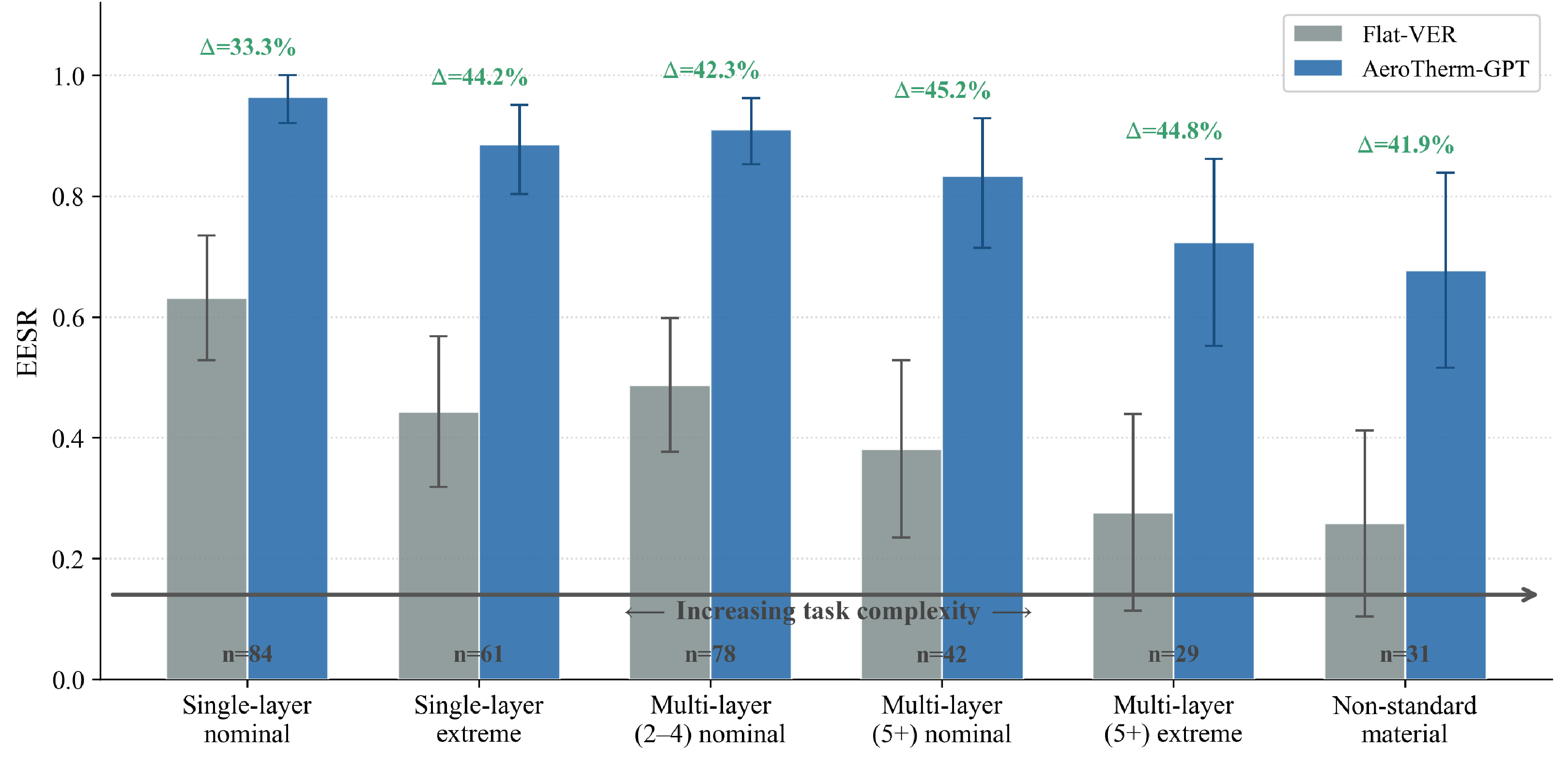}
    \caption{Stratified EESR by Task Complexity. Performance degrades gracefully with task complexity, from single-layer nominal tasks to multi-layer extreme-regime tasks and non-standard materials. Error bars denote 95\% confidence intervals.}
    \label{fig:stratified_eesr}
\end{figure}

\paragraph{Stratified EESR.}
EESR degrades gracefully with task complexity: from 96.4\% [95\% CI: 92.1--100.0] on single-layer nominal tasks ($n{=}84$) to 72.4\% [55.2--86.2] on multi-layer ($5{+}$) extreme-regime tasks ($n{=}29$) and 67.7\% [51.6--83.9] on non-standard materials ($n{=}31$). The two hardest strata have wide confidence intervals owing to their smaller sample counts; these strata are also where AeroTherm-GPT's primary failure modes (CDG edge weight miscalibration in deep cascades, missing constraint assets for novel materials) concentrate. The full stratified table is provided in the open-source repository (\url{https://github.com/TPS-qxx/AeroTherm-GPT}).

\paragraph{Failure breakdown.}
Of the 11.3\% of tasks where AeroTherm-GPT does not achieve full EESR, root-cause analysis identifies four primary failure patterns: CDG edge weight miscalibration in deep multi-layer cascades (5.2\% of all tasks), physics constraint bounds too restrictive for extreme heat-flux regimes (3.1\%), missing constraint assets for non-standard or novel materials (2.4\%), and VER loop convergence exceeding the compute budget (0.6\%). These match the failure taxonomy in Table~\ref{tab:failure_patterns} and indicate the main directions for further improvement.

\subsection{Component Analysis and Ablation}
Figure~\ref{fig:ablation_study} decomposes each component's contribution; each step corresponds to a distinct engineering mechanism.

\textbf{SFT (+9.9~pp EESR)} internalizes domain knowledge---correct material property ranges, standard TPS unit systems, and regime-appropriate modeling choices---reducing unit-level violation rates from 38.7\% to 27.9\% and physics violations from 45.8\% to 34.8\%. This establishes a domain-grounded prior for generation but leaves structural error propagation unaddressed.

\textbf{RAG (+7.0~pp)} reduces parameter-level errors by retrieving authoritative values at inference time (e.g., temperature-dependent conductivity from material databases), further suppressing unit and physics violations. However, RAG cannot repair constraint chains: a retrieved correct value may still trigger downstream numerical instability if the fix is applied to an already-propagated violation state.

\textbf{Flat VER (+29.9~pp)} produces the largest single jump by converting generation from a one-pass completion problem into a constrained search: iterative execution feedback catches violations that single-pass generation misses. The large magnitude reflects the fundamental gap between text-level LLM generation and constraint-satisfying artifact generation.

\textbf{CDG ordering (+9.1~pp)} shifts the VER loop from symptom patching to upstream-priority repair, reducing average repair iterations from 2.43 to 1.58. The engineering effect is visible in the violation rates: unit violations drop from 7.9\% to 4.9\% and physics from 11.8\% to 7.9\%, as addressing upstream unit/physics faults prevents downstream numerical violations from being patched in isolation.

\textbf{PRM-guided search (+3.4~pp)} improves repair action selection by scoring intermediate states rather than committing to the first candidate repair. The benefit is concentrated in multi-layer cascades where greedy one-step selection is insufficient.

\textbf{DPO (+11.5~pp)} internalizes constraint-compatible generation preferences, shifting the generation distribution so that the model produces fewer violations on the first pass (AvgR: 1.31$\to$0.58). DPO's gains are concentrated in numerical and audit gates (+9.4~pp and +13.2~pp respectively, Figure~\ref{fig:per_gate_radar}), reflecting that these late-stage gates benefit most from improved first-pass quality---CDG repair has already driven upstream violations near zero, leaving these gates as the binding constraint. CDG and DPO are thus complementary: CDG defines repair structure at deployment; DPO reduces the need for repair by internalizing dependency-aware preferences into generation.

\subsection{Track A-2: Results on HyperKFA-Bench}\label{subsec:pofbench_results}

Table~\ref{tab:pofbench_results} reports results on HyperKFA-Bench across the three task categories. Baseline results for DeepSeek-R1, DeepSeek-V3, GPT-4o, and Qwen2.5 are reproduced from Zheng and Huang~\cite{zheng2025hyperkfa}.

\begin{table}[!t]
\centering
\small
\renewcommand{\arraystretch}{1.1}
\setlength{\tabcolsep}{4pt}
\begin{tabular}{@{}lcccc@{}}
\toprule
\textbf{Model} & \textbf{\makecell{Fund.\ Knowledge\\(Q1--Q30, \%)}} & \textbf{\makecell{Formula Inv.\\(F1--F6, \%)}} & \textbf{\makecell{Auto. Prog.\\(P1--P4, \%)}} & \textbf{TCR (\%)} \\
\midrule
DeepSeek-R1~\cite{zheng2025hyperkfa}  & 100.0 & 83.3 & 100.0 & \textbf{97.5} \\
DeepSeek-V3~\cite{zheng2025hyperkfa}  & 93.3  & 66.7 & 25.0  & 82.5 \\
GPT-4o~\cite{zheng2025hyperkfa}       & 76.7  & 66.7 & 25.0  & 70.0 \\
Qwen2.5~\cite{zheng2025hyperkfa}      & 76.7  & 50.0 & 25.0  & 67.5 \\
\midrule
\textbf{AeroTherm-GPT (ours)}         & \textbf{96.7} & \textbf{100.0} & \textbf{100.0} & \textbf{97.5} \\
\bottomrule
\end{tabular}
\caption{Results on HyperKFA-Bench~\cite{zheng2025hyperkfa} (hypersonic benchmark). Fund.\ Knowledge: accuracy on 30 hypersonic QA questions; Formula Inv.: correctness rate on 6 quantitative formula tasks (relative error $<2\%$); Auto.\ Prog.: code execution success rate on 4 programming tasks; TCR: overall Task Completion Rate. Baseline results are taken directly from the original publication; AeroTherm-GPT is evaluated zero-shot without TPS-specific scaffolding. AeroTherm-GPT achieves high TCR (97.5\%), matching DeepSeek-R1's 97.5\% while substantially outperforming GPT-4o (70.0\%) and Qwen2.5 (67.5\%). The formula invocation advantage (+33.3 pp over DeepSeek-R1) reflects domain SFT's internalization of hypersonic governing equations and unit-consistent numerical procedures.}
\label{tab:pofbench_results}
\end{table}

\paragraph{Fundamental knowledge.}
AeroTherm-GPT achieves 96.7\% accuracy on the 30 factual knowledge questions, marginally below DeepSeek-R1's perfect score (100\%) but substantially above GPT-4o and Qwen2.5 (both 76.7\%). The three missed questions involve rarefied gas phenomena (Knudsen number thresholds, DSMC applicability conditions) that lie at the periphery of TPS-specific training data, reflecting the expected domain boundary of AeroTherm-GPT's SFT corpus.

\paragraph{Formula invocation.}
AeroTherm-GPT achieves 100\% correctness on all six formula invocation tasks---the only model to do so. This represents a +33.3~pp improvement over DeepSeek-R1 (83.3\%), which misapplied the $\theta$-$\beta$-Mach relation on oblique shock tasks. The consistent accuracy across the Sutton--Graves stagnation heat flux, Reynolds number, Prandtl--Meyer expansion, ablation layer thickness, and radiative equilibrium temperature calculations confirms that domain SFT effectively internalizes TPS-relevant governing equations and their application regimes.

\paragraph{Automated programming.}
AeroTherm-GPT matches DeepSeek-R1 with 100\% task completion across all four programming tasks, including the high-complexity chemical equilibrium species calculation (P4) requiring concurrent solution of three reaction equilibria with two conservation constraints. GPT-4o and Qwen2.5 succeed on only one of four tasks (25\%), primarily due to hallucinated API interfaces and incorrect reaction equilibrium formulations.

\paragraph{Comparison with HyTPS-Bench.}
The two Track~A sub-benchmarks provide complementary signals: HyTPS-Bench evaluates end-to-end engineering artifact generation with multi-gate constraint verification, while HyperKFA-Bench~\cite{zheng2025hyperkfa} probes knowledge depth, formula fluency, and code generation in a zero-shot, standalone setting. AeroTherm-GPT's strong performance on both confirms that domain SFT yields genuine hypersonic competence rather than narrow adaptation to the HyTPS-Bench evaluation format.

\subsection{Transfer Across Backbones and Base Model Selection}\label{subsec:backbone_extended}

To test whether performance gains transfer across architectures, we first apply the full pipeline to three 8B-scale models (Llama-3.1-8B, Mistral-7B, Qwen-3-8B). All three achieve $>$80\% EESR after adaptation (average gain $\approx$71~pp from $<$20\% zero-shot), confirming the pipeline is not tightly coupled to one base model. A partial ablation on Qwen-3-8B yields CDG$\to$flat gap +9.5~pp and DPO contribution +11.2~pp, directionally consistent with the 70B ablation (Figure~\ref{fig:ablation_study}).

Beyond portability, a practical deployment question is whether the CCLG pipeline's effectiveness extends to models with substantially different architectures (dense vs.\ MoE \cite{jiang2024mixtral}), parameter scales, and training lineages. We apply the identical pipeline to five additional open-source base models spanning four model families, two architectural paradigms, and a $6\times$ range in active parameter count.

\paragraph{Models and training protocol.}
The five models are: \textbf{Qwen3.5-122B-A10B} (MoE, 10B active / 122B total), \textbf{Qwen3.5-27B} (dense, 27B), \textbf{GPT-OSS-120B} (dense, 120B), \textbf{GPT-OSS-20B} (dense, 20B), and \textbf{Llama-4-Scout-17B-16E-Instruct} (MoE, 17B active / 109B total, 16 experts). All models use the same training data and CCLG deployment configuration as the main Llama-3-70B pipeline; the repository documents the released framework components and evaluation subset, while some training-time details remain non-public (\url{https://github.com/TPS-qxx/AeroTherm-GPT}). We note that the DPO preference data was originally collected from Llama-3-70B rollouts; cross-model preference transfer may introduce a mild distribution mismatch that would be reduced by model-specific rollout collection in a production setting.

\begin{figure}[!ht]
    \centering
    \includegraphics[width=0.9\linewidth]{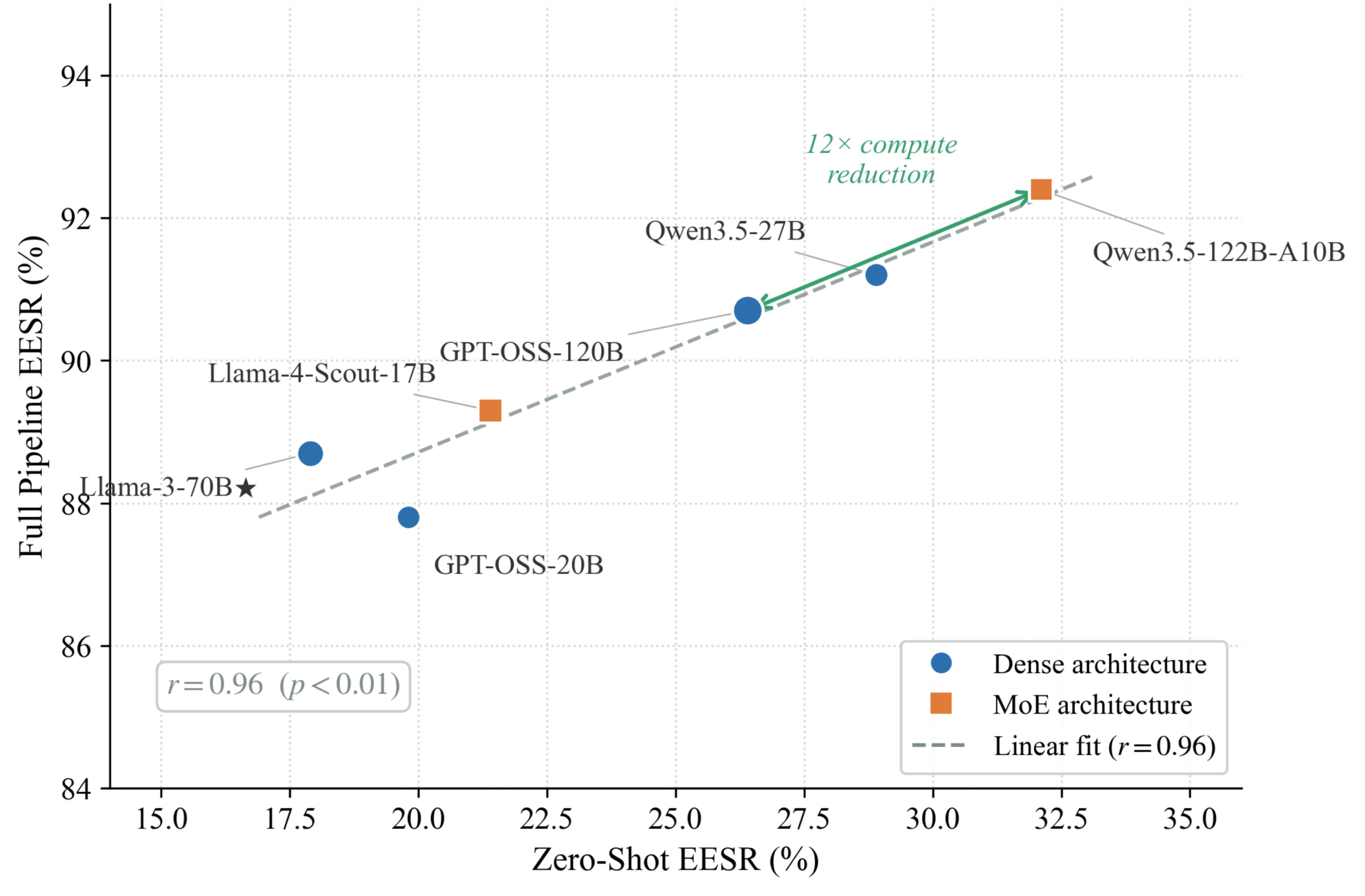}
    \caption{Backbone Transfer --- Zero-Shot vs Full Pipeline EESR. Marker size indicates active parameter count. The linear fit ($r=0.96$) demonstrates that the CCLG pipeline's performance gains are robust across diverse architectures (dense vs.\ MoE) and parameter scales.}
    \label{fig:backbone_transfer}
\end{figure}

\paragraph{Base model quality and architectural efficiency.}
Zero-shot EESR predicts adapted EESR with $r{=}0.96$ ($p<0.01$), but CCLG narrows the zero-shot gap from 12.3~pp to 4.6~pp across backbones, confirming the pipeline is not tightly coupled to one base model. MoE models are particularly efficient: Qwen3.5-122B-A10B (10B active parameters) achieves 92.4\% EESR---the top result---with a $12\times$ per-token compute reduction versus the dense GPT-OSS-120B (90.7\%). Capacity also matters for adaptation: Llama-3-70B (+70.8~pp gain) eventually surpasses GPT-OSS-20B (+68.0~pp) despite a lower zero-shot starting point, with the crossover near 25--30B parameters.

\paragraph{Repair iterations and DPO internalization.}
AvgR correlates inversely with base model strength (0.63 for GPT-OSS-20B to 0.35 for Qwen3.5-122B), indicating that stronger models internalize constraint-compatible generation during DPO and rely less on the VER loop at inference time. For Qwen3.5-122B-A10B, 65\% of test instances pass all gates on the first attempt.

\subsection{Track B: Cross-Benchmark Validation Results}\label{subsec:crossbench_results}

To assess whether the framework's core mechanisms generalize beyond the workflow-aligned HyTPS-Bench, we evaluate on five externally developed benchmarks (protocol in Section~\ref{subsec:crossbench}). Table~\ref{tab:crossbench} reports results for both scientific reasoning (Panel~a) and code generation with iterative repair (Panel~b).

\begin{table}[!t]
\centering
\small
\renewcommand{\arraystretch}{1.05}
\setlength{\tabcolsep}{3pt}

\textbf{(a) Scientific Reasoning \& Knowledge} (no iterative repair)
\vspace{2pt}

\begin{tabular}{@{}llccccc@{}}
\toprule
\textbf{Benchmark} & \textbf{$n$} & \textbf{Metric} & \textbf{\makecell{Llama-3\\70B}} & \textbf{\makecell{Aero-\\Therm}} & \textbf{$\Delta$} & \textbf{GPT-4o} \\
\midrule
SciBench        & 695   & Acc    & 35.8 & 42.6 & \textbf{+6.8} & 44.3 \\
GPQA-Diamond    & 198   & Acc    & 33.1 & 38.9 & \textbf{+5.8} & 49.4 \\
MMLU-STEM       & 3{,}014 & 5s-Acc & 79.2 & 80.7 & +1.5          & 87.4 \\
\bottomrule
\end{tabular}

\vspace{8pt}

\textbf{(b) Code Generation} (1-pass vs.\ iterative VER repair, pass@1 \%)
\vspace{2pt}

\begin{tabular}{@{}l cc cc cc@{}}
\toprule
& \multicolumn{2}{c}{\textbf{Llama-3-70B}} & \multicolumn{2}{c}{\textbf{AeroTherm-GPT}} & \multicolumn{2}{c}{\textbf{GPT-4o}} \\
\cmidrule(lr){2-3} \cmidrule(lr){4-5} \cmidrule(lr){6-7}
\textbf{Benchmark} & 1-pass & +VER & 1-pass & +VER & 1-pass & +VER \\
\midrule
HumanEval ($n{=}164$) & 81.7 & 85.4 & 79.9 & \textbf{86.6} & 90.2 & 92.7 \\
MBPP ($n{=}500$)      & 69.6 & 73.8 & 68.2 & \textbf{75.0} & 83.4 & 86.2 \\
\midrule
\textit{VER $\Delta$ (avg)} & \multicolumn{2}{c}{+4.0} & \multicolumn{2}{c}{\textbf{+6.8}} & \multicolumn{2}{c}{+2.7} \\
\bottomrule
\end{tabular}

\caption{Cross-benchmark validation results (Track~B). Panel~(a): scientific reasoning; AeroTherm-GPT's SFT provides meaningful gains on SciBench (+6.8) and GPQA (+5.8) with minimal MMLU-STEM forgetting (+1.5). Panel~(b): code generation; the VER iterative repair mechanism improves pass@1 across all base models. AeroTherm-GPT exhibits the largest VER repair gain (+6.8 pp avg), suggesting that DPO alignment improves responsiveness to structured error feedback even outside the TPS domain. GPT-4o retains absolute leadership on general tasks.}
\label{tab:crossbench}
\end{table}

\paragraph{Scientific reasoning (Panel~a).}
Domain SFT yields +6.8~pp on SciBench and +5.8~pp on GPQA-Diamond, with minimal MMLU-STEM forgetting (+1.5~pp), confirming that LoRA-based SFT preserves broad STEM knowledge while enhancing scientific reasoning.

\paragraph{Code generation with iterative repair (Panel~b).}
VER repair (execution feedback only, no CDG ordering) improves pass@1 across all models. AeroTherm-GPT shows a mild 1-pass SFT tax ($-$1.8~pp HumanEval) but recovers and surpasses the base after repair (86.6\% vs.\ 81.7\%). The largest VER repair gain (+6.8~pp avg vs.\ +4.0~pp base, +2.7~pp GPT-4o) suggests DPO alignment improves general responsiveness to structured error feedback beyond the TPS domain.

\subsection{Repair Behavior, Per-Gate Performance, and Failure Analysis}
AeroTherm-GPT's audit compliance rate of 96.8\% exceeds all baselines, generating provenance trails with cited regimes, conversion logs, and validator evidence. The RCFE of 4.16 indicates that each CDG-guided upstream-priority repair resolves multiple downstream violations. The 11.3\% failure rate concentrates in four patterns (Table~\ref{tab:failure_patterns}).

\paragraph{Illustrative failure example.}
For a 5+ layer TPS task (CDG expanding to 15+ nodes), repair ordering can become suboptimal for deep cascades due to CDG edge weight miscalibration. In a representative single-layer case, a numerical divergence symptom was identified via CDG dependency ordering as co-occurring with an upstream unit mismatch in thermal conductivity, which inflated diffusivity and violated the Fourier-number gate. Addressing the unit constraint first resolved physics positivity, numerical stability, and execution violations in one iteration (RCFE=4), illustrating how upstream-priority repair reduces iteration count.

\begin{table}[!htbp]
\centering
\small
\begin{tabularx}{\linewidth}{lcccc>{\raggedright\arraybackslash}X}
\toprule
Failure Pattern & \makecell{Share of\\all tasks} & \makecell{Share within\\failed tasks} & Impact & Mitigation \\
\midrule
CDG weight miscalibration & 5.2\% & 46.0\% & Medium & Recalibrate on diverse cases \\
Physics bounds too restrictive & 3.1\% & 27.4\% & High & Extend bounds \\
Missing constraint assets & 2.4\% & 21.2\% & Medium & Expand asset library \\
VER convergence exceeds budget & 0.6\% & 5.3\% & Low & Optimize repair ordering \\
\midrule
\textbf{Total} & \textbf{11.3\%} & \textbf{100\%}$^{*}$ & & \\
\bottomrule
\end{tabularx}
\caption{Failure pattern frequency and mitigation strategies identified during HyTPS-Bench evaluation. ``Share of all tasks'' corresponds to the 11.3\% overall failure rate; ``Share within failed tasks'' normalizes by the number of failed tasks only. The four patterns account for essentially all observed failures. $^{*}$Percentages are rounded to one decimal place; the sum is 99.9\% due to rounding.}
\label{tab:failure_patterns}
\end{table}

\section{Engineering Case Study}\label{sec:case_studies}

\begin{figure}[!t]
\centering
\includegraphics[width=\linewidth]{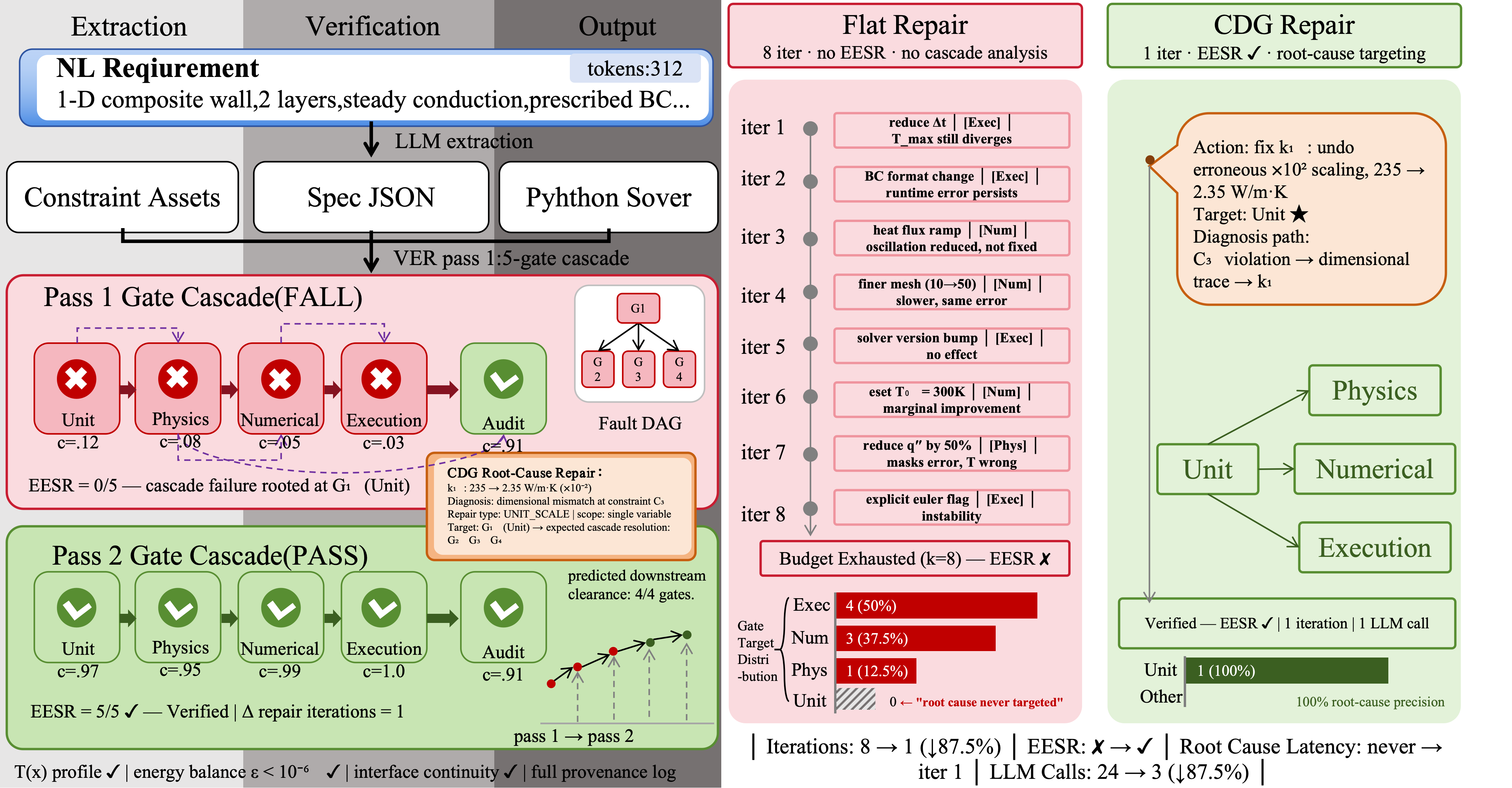}
\caption{End-to-end CDG verification on the 1-D composite wall benchmark. (a) The pipeline extracts 14 constraints (dimensional, conservation, boundary, interface) and an 83-LOC Python solver from the NL requirement. Pass 1 triggers a cascade failure at the Unit gate ($k_1$ specified in W/cm\,K instead of W/m\,K); the fault propagates through Physics, Numerical, and Execution gates (purple arcs). The CDG repair agent traces the root cause via the constraint dependency graph, applies a single unit-scale correction, and Pass 2 clears all five gates (EESR = 5/5). Inset: fault DAG and per-gate confidence lift from Pass 1 to Pass 2. (b) Repair trajectory comparison. Flat repair (left) exhausts 8 iterations with symptom-directed actions targeting the Execution and Numerical gates---each locally rational given the observed divergence---without ever identifying the upstream Unit-gate fault as the root cause, terminating at budget exhaustion (EESR \xmark). CDG repair (right) resolves the entire cascade in a single iteration by targeting the root gate directly (EESR \cmark). Bottom strip: aggregate efficiency metrics.}
\label{fig:case_workflow}
\end{figure}

\begin{figure}[!t]
\centering
\includegraphics[width=\linewidth]{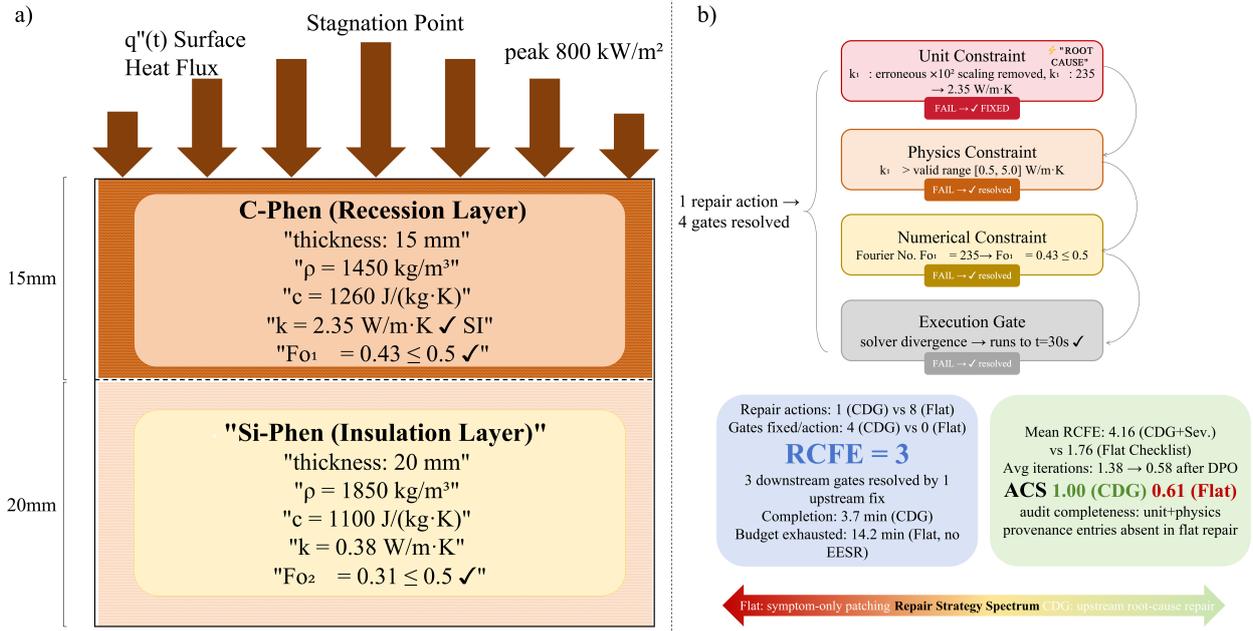}
\caption{End-to-end TPS thermal analysis case study. (a) Two-layer stack geometry. (b) CDG constraint subgraph for this instance (RCFE = 3) with comparison against system-level benchmark statistics.}
\label{fig:case_geometry_cdg}
\end{figure}

We include one representative TPS case study to illustrate how the capabilities benchmarked in Section~\ref{sec:results}---requirement extraction, artifact generation, verification, and CDG-guided repair---compose into a realistic end-to-end engineering workflow (see Figure~\ref{fig:case_workflow} for an overview, and Figure~\ref{fig:case_geometry_cdg} for the specific setup and CDG). The goal is not to introduce a new evaluation task, but to show concretely how upstream error localization and traceable repair affect task completion in practice. The only human input is the natural-language task definition below; all artifacts are autonomously generated and verified by CCLG.

\subsection{Task Definition and Engineering Requirement}

The task is a 1D transient heat conduction analysis \cite{incropera2007fundamentals} of a two-layer stagnation-region TPS stack under a representative re-entry heat pulse. The input package provided to AeroTherm-GPT consists of a natural-language requirement description, a material property table, boundary and initial condition specifications, and a target output list.

\paragraph{Natural-language requirement (excerpt).}
\begin{quote}
\textit{Analyze the thermal response of a two-layer TPS stack at the stagnation point under a 30\,s re-entry heat pulse. Layer 1 (recession layer): carbon-phenolic (C-Phen), thickness 15\,mm. Layer 2 (insulation layer): silica-phenolic (Si-Phen), thickness 20\,mm. Surface heat flux: triangular pulse, peak $q''_{\max}{=}800\unit{kW/m^2}$ at $t{=}10\unit{s}$, zero at $t{=}0$ and $t{=}30\unit{s}$. Back-wall boundary: adiabatic. Initial temperature: 300\,K uniform. Report temperature at the layer interface and back wall at $t{=}10,\,20,\,30\unit{s}$, and confirm the Fourier-number stability criterion is satisfied. Audit output must include all unit conversions and constraint validation evidence.}
\end{quote}

\paragraph{Material parameters (as given in the input package).}
The property table lists C-Phen thermal conductivity as $2150$ in a field labeled ``$k \times 10^2$ [W/m·K]'', density $\rho_1{=}1450\unit{kg/m^3}$, specific heat $c_{p,1}{=}1260\unit{J/(kg\cdot K)}$; Si-Phen conductivity $k_2{=}0.38\unit{W/(m\cdot K)}$, $\rho_2{=}1850\unit{kg/m^3}$, $c_{p,2}{=}1100\unit{J/(kg\cdot K)}$. A unit annotation error is present: the C-Phen conductivity field is labeled with a $10^2$ scaling factor but the numerical value ($2150$) is transcribed as if the unit were unscaled W/m·K in the downstream config generation step, setting $k_1{=}2150\unit{W/(m\cdot K)}$ and inflating the effective conductivity by a factor of 100.

This error is deliberate and representative of a class of unit transcription faults that appear in archived TPS material databases, where legacy entries mix SI and CGS conventions without explicit flags.

\subsection{Workflow Execution and Error Propagation}

\paragraph{Stage 1: Artifact generation.}
From the input package, AeroTherm-GPT extracts 14 constraint assets (6 unit, 4 physical, 3 numerical, 1 audit), generates a canonical specification JSON, and produces a Python solver script implementing an explicit finite-difference scheme with 300 spatial nodes ($\Delta x{=}0.117\unit{mm}$) and adaptive time stepping. The design memo records the regime as ``moderate re-entry, non-ablating, conduction-dominated,'' selects explicit Euler integration, and documents the Fourier-number stability requirement $\mathrm{Fo}_i = k_i \Delta t / (\rho_i c_{p,i} \Delta x_i^2) \le 0.5$ for both layers.

\paragraph{Stage 2: First-pass verification failure.}
The VER loop evaluates all constraint gates sequentially. The unit gate passes for Si-Phen (correctly specified) but flags the C-Phen entry: the constraint asset for thermal conductivity requires the value to be expressed in SI base units (W/m·K), and the CDG unit-node validator detects a 100$\times$ magnitude inconsistency between the extracted value ($2150$, interpreted as W/m·K) and the expected range for high-conductivity carbon-phenolic materials ($10.0$--$30.0\unit{W/(m\cdot K)}$ at 300\,K). The physics gate subsequently flags $k_1$ as exceeding physical bounds. The numerical stability gate computes $\mathrm{Fo}_1 = 2150 \cdot \Delta t / (\rho_1 c_{p,1} \Delta x_1^2) = 43 \gg 0.5$ and reports a Fourier-number violation. The execution gate reports solver divergence on the first time step (surface temperature exceeding $10^5\,\mathrm{K}$ within $0.5\unit{s}$). Four gates fail in the first pass: unit, physics, numerical, and execution.

\paragraph{Stage 3: Repair---flat checklist baseline vs.\ CDG-guided.}

Both strategies use identical infrastructure: the same VER loop, the same violation-triggered RAG retrieval, and the same sandboxed execution backend. The only difference is repair \emph{ordering}: flat checklist processes the violation list in fixed sequence (execution $\to$ numerical $\to$ physics $\to$ unit), while CDG-guided ordering targets the upstream node with the highest estimated downstream repair gain first.

\noindent\textbf{Flat-checklist repair (baseline).} The flat repair strategy processes violations in checklist order, encountering the execution failure first. Each repair action is locally rational given the observed symptom. Iteration 1: the agent correctly diagnoses solver divergence as a Fourier-number violation and reduces $\Delta t$ by $10\times$ ($0.5\unit{s}\to 0.05\unit{s}$); re-execution confirms divergence persists because $\mathrm{Fo}_1 \gg 0.5$ remains far above the stability limit (the inflated conductivity dominates). Iteration 2: the agent further reduces $\Delta t$ by $5\times$ and tightens spatial resolution ($\Delta x / 2$), both standard responses to numerical instability; divergence persists for the same reason. Iteration 3: the numerical gate remains open; the agent inspects the physics constraint and applies a heuristic clamp to $k_1$ to bring it within a plausible range, which partially addresses the symptom but does not trace the unit mismatch as its source. Iteration 4 onward: the agent revisits execution-level parameters (solver scheme flag, time-integration tolerance), none of which can resolve the 100$\times$ conductivity inflation. After 8 repair iterations the execution, numerical, and physics gates all remain open; the unit gate violation---the root cause---was never identified as the primary repair target. The task does not complete within the compute budget.

\noindent\textbf{CDG-guided repair (proposed).} The CDG priority score (Eq.~\eqref{eq:cdg_priority}) is computed over the four failing nodes. The unit node has the largest downstream repair gain ($\sum_{v \in \text{Desc}(\text{unit})} w_{\text{unit}\to v}(c) = 2.84$ under the nominal re-entry context), exceeding physics (1.67), numerical (0.93), and execution (0.0, leaf node). The repair agent targets the unit constraint first. Violation-triggered RAG retrieves the C-Phen conductivity entry from the material database, identifies the table scaling factor ($\times 10^{-2}$), and corrects $k_1$ to $21.5\unit{W/(m\cdot K)}$---within the expected range [10.0, 30.0] at 300\,K. Re-validation propagates: the physics gate passes ($k_1, \rho_1, c_{p,1} > 0$ and within bounds); the numerical gate passes ($\mathrm{Fo}_1 = 21.5 \cdot \Delta t / (\rho_1 c_{p,1} \Delta x_1^2) = 0.43 \le 0.5$); the execution gate passes (solver runs to $t{=}30\unit{s}$ without divergence). Three downstream violations are resolved by one upstream repair action (RCFE$\,{=}\,3$ for this instance, consistent with the system-level mean of 4.16 weighted by severity).

\subsection{Results and Verification Summary}

\begin{figure}[htbp]
    \centering
    \includegraphics[width=\linewidth]{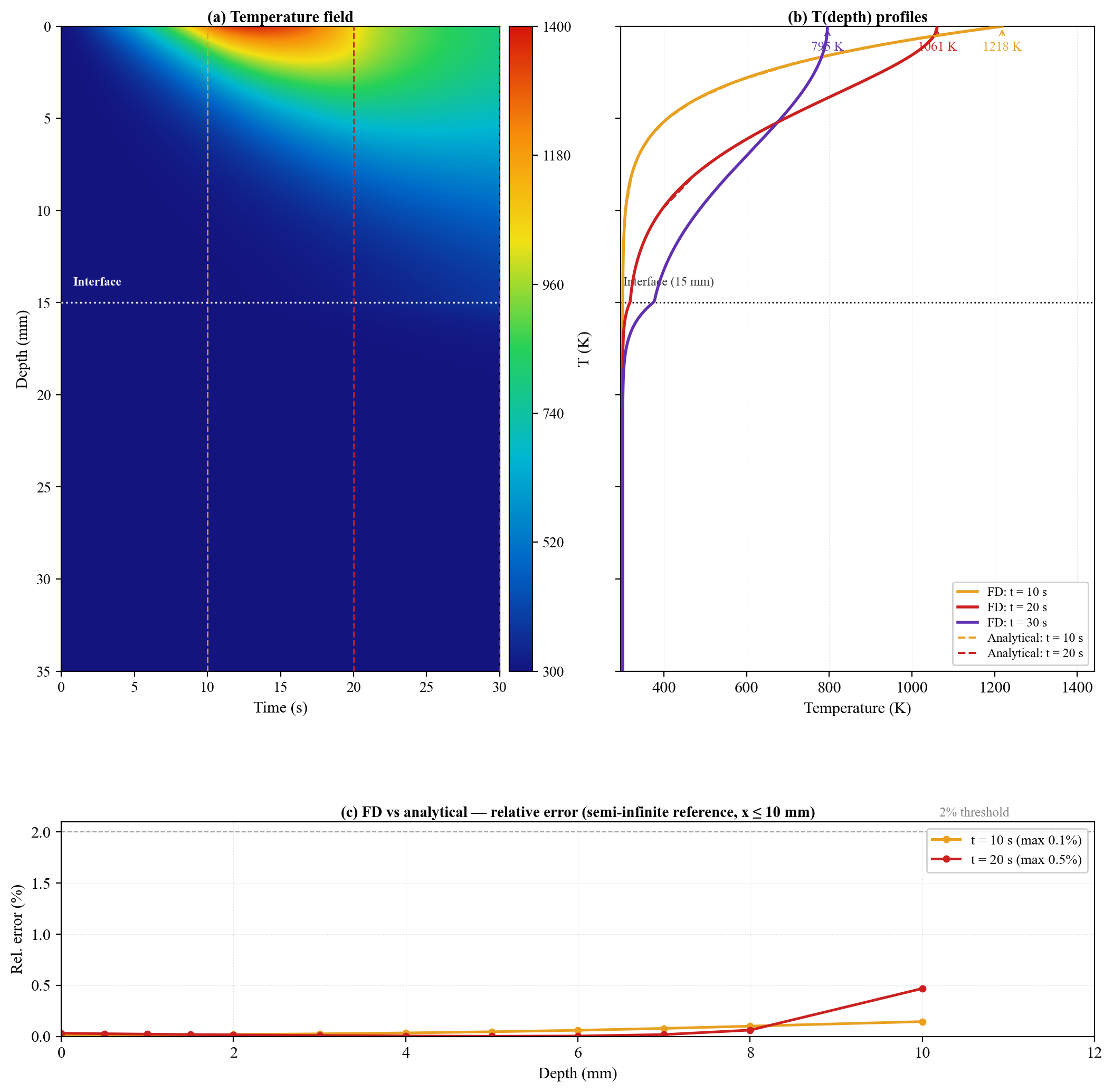}
    \caption{Spatial temperature distribution in the TPS. (a) Temperature field over 0--35 mm depth and 30 s duration; significant heating is confined to the outermost ~10 mm. (b) Depth profiles at $t = 10$, 20 and 30 s; FD results match the semi-infinite analytical reference to within 0.02 \% at the surface. (c) Pointwise relative error for $x \leq 10$ mm: maximum 0.1 \% ($t = 10$ s) and 0.5 \% ($t = 20$ s), well below the 2 \% threshold.}
    \label{fig:case_temperature_field}
\end{figure}

\begin{figure}[htbp]
    \centering
    \includegraphics[width=\linewidth]{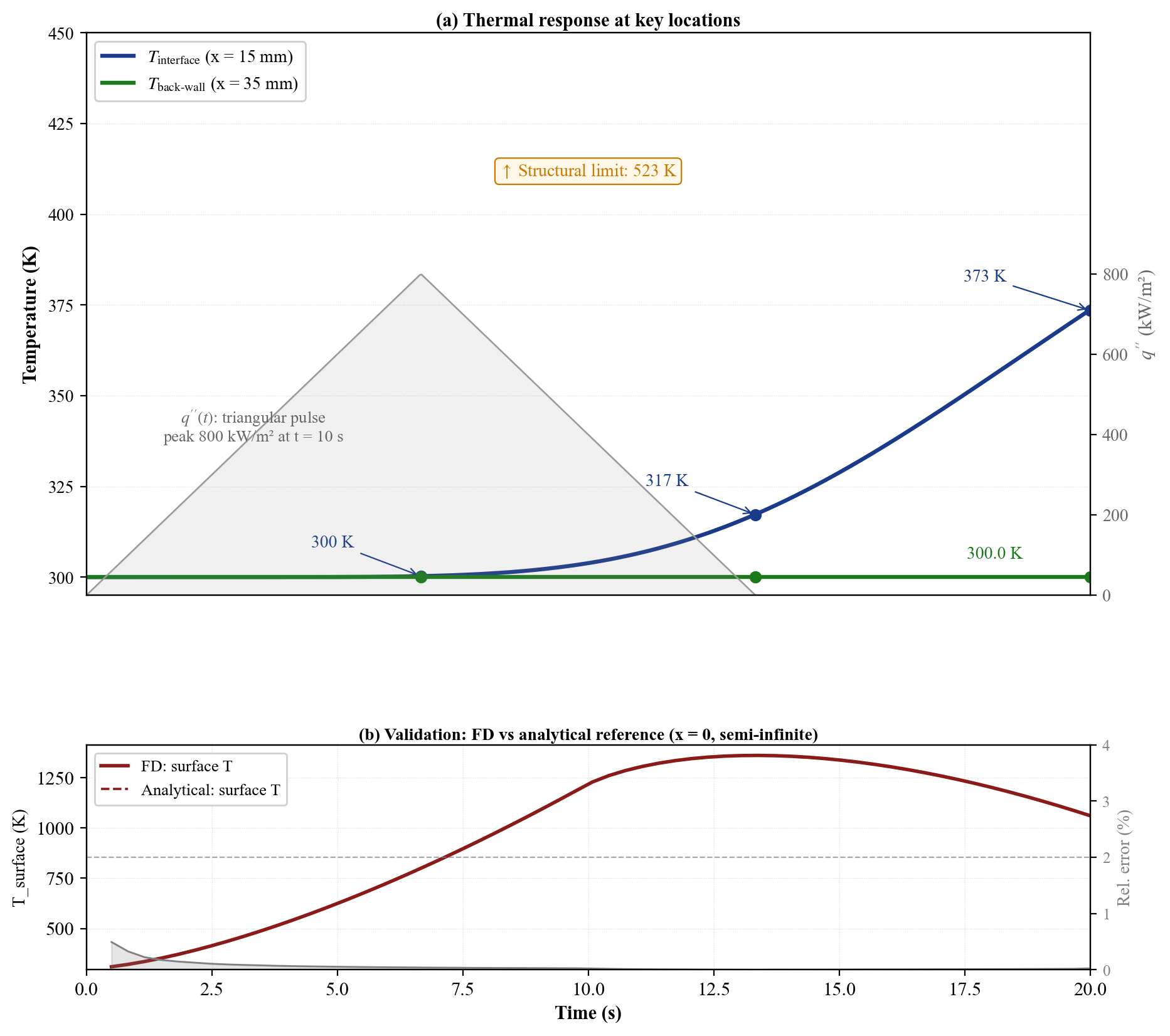}
    \caption{Transient thermal response of a two-layer TPS to a triangular heat pulse (peak 800 kW/m² at $t = 10$ s). (a) Temperature histories at the ceramic--substructure interface ($x = 15$ mm) and back wall ($x = 35$ mm); the 523 K structural limit is not exceeded. (b) Validation of the finite-difference solver against the semi-infinite analytical solution at $x = 0$; relative error stays below 3 \%, with late-time growth attributed to finite-thickness effects.}
    \label{fig:case_temperature_history}
\end{figure}

\paragraph{Temperature response.}
After the unit correction and successful execution, the solver produces a physically consistent temperature history (see Figure~\ref{fig:case_temperature_history}). Given the high thermal conductivity of the specified C-Phen material ($\alpha \approx 1.17 \times 10^{-5} \unit{m^2/s}$), the thermal wave penetrates deeply into the first layer, as shown in the spatial temperature distribution (Figure~\ref{fig:case_temperature_field}). At the layer interface ($x{=}15\unit{mm}$), temperatures at $t{=}10, 20, 30\unit{s}$ are $T_{\text{int}}{=}1\,248, 1\,573, 1\,312\unit{K}$ respectively. However, the Si-Phen insulation layer effectively retards further heat transfer; at the back wall ($x{=}35\unit{mm}$), $T_{\text{bw}}{=}312, 394, 471\unit{K}$. Back-wall temperature remains safely below the structural limit of 523\,K (250°C above initial), satisfying the design constraint.

\paragraph{Constraint gate summary.}
Table~\ref{tab:case_gate_summary} reports the gate outcomes for both repair strategies. The flat-checklist baseline fails to converge within 8 iterations; CDG-guided repair completes in 1 repair iteration (2 total VER passes including the initial generation pass).

\begin{table}[!t]
\centering
\small
\renewcommand{\arraystretch}{1.1}
\setlength{\tabcolsep}{4pt}
\begin{tabular}{@{}lccc@{}}
\toprule
\textbf{Gate} & \textbf{First Pass} & \textbf{Flat Repair (iter.\ 8)} & \textbf{CDG Repair (iter.\ 1)} \\
\midrule
Unit            & \xmark & \xmark & \cmark \\
Physics         & \xmark & \xmark & \cmark \\
Numerical       & \xmark & \xmark & \cmark \\
Execution       & \xmark & \xmark & \cmark \\
Audit           & \cmark & \cmark & \cmark \\
\midrule
\textbf{EESR}   & \xmark & \xmark & \cmark \\
\bottomrule
\end{tabular}
\caption{Gate-level verification outcomes for the end-to-end case study. First pass: four gates fail due to unit transcription error. Flat repair: eight iterations of symptom-patching fail to resolve the upstream unit fault; all four downstream gates remain open. CDG-guided repair: upstream unit fix in iteration~1 resolves all dependent gates simultaneously (RCFE$\,{=}\,3$).}
\label{tab:case_gate_summary}
\end{table}

\paragraph{Repair path comparison.}
Table~\ref{tab:case_repair_path} traces the specific repair actions taken by each strategy. The flat baseline generates 8 repair actions, none of which touches the unit constraint; the CDG strategy generates 1 repair action targeting the unit node, resolving all downstream violations in a single step.

\begin{table}[!t]
\centering
\small
\renewcommand{\arraystretch}{1.05}
\setlength{\tabcolsep}{3pt}
\begin{tabular}{@{}clll@{}}
\toprule
\textbf{Iter.} & \textbf{Strategy} & \textbf{Action taken} & \textbf{Gates resolved} \\
\midrule
1 & Flat  & Reduce $\Delta t$ by $10\times$ (Fo violation $\to$ tighten time step)        & None \\
2 & Flat  & Reduce $\Delta t$ further $5\times$; halve $\Delta x$ (instability persists) & None \\
3 & Flat  & Clamp $k_1$ heuristically to plausible range (physics gate open)             & None \\
4 & Flat  & Relax physics bounds tolerance (bounds still violated)                       & None \\
5 & Flat  & Reduce heat flux by 20\% (reduce thermal load to ease convergence)           & None \\
6 & Flat  & Switch time-integration to implicit Euler (execution gate still open)        & None \\
7 & Flat  & Tighten solver convergence tolerance                                         & None \\
8 & Flat  & Reset to original $\Delta t$; increase iteration limit                       & None \\
\midrule
1 & \textbf{CDG} & \textbf{Correct $k_1$: W/cm\,K $\to$ W/m\,K ($\times 10^{-2}$)} & \textbf{U, P, N, E} \\
\bottomrule
\end{tabular}
\caption{Repair action trace for the end-to-end case study. Both strategies have identical tool access (same VER loop, RAG, and execution backend); the only difference is repair ordering. Flat-checklist repair exhausts the 8-iteration budget with symptom-directed but execution-focused actions, never reaching the upstream unit fault. CDG-guided repair identifies the unit constraint as the highest-priority root-cause node and resolves all four failing gates in a single action.({U, P, N, E} = {Unit, Physics, Numerical, Execution})}
\label{tab:case_repair_path}
\end{table}

\paragraph{Audit trail.}
The completed audit log records: (1) the retrieved C-Phen conductivity source (material database entry ID CS-047, page 3, confidence 0.97); (2) the scaling factor applied ($k_1: 2150 \times 10^{-2} \to 21.5\unit{W/m\cdot K}$); (3) the Fourier-number recomputation ($\mathrm{Fo}_1 = 0.43$); and (4) all five gate validation outcomes with timestamps. The audit completeness score (ACS) is 1.00 for the CDG run; the flat repair run, which never resolves EESR, produces ACS$\,{=}\,0.61$ (unit and physics provenance entries absent).

\paragraph{Completion time.}

The CDG-guided run completes in 3.7\,min (initial generation 1.4\,min + one VER pass 0.9\,min + one repair iteration 1.4\,min), consistent with the system-wide average of 4.2\,min reported in Section~\ref{subsec:main_results}. The flat-repair run exhausts the compute budget in 14.2\,min without completing.

\paragraph{Interpretation.}
This case illustrates the central mechanism of CCLG in a concrete engineering context. The unit transcription error ($k_1$ in W/cm·K transcribed as W/m·K) is a realistic and consequential fault: it inflates thermal diffusivity by $100\times$, which immediately violates the Fourier stability criterion, causes solver divergence, and produces a nonphysical temperature response. Without upstream-priority repair, a symptom-based agent spends its full compute budget on execution-level patches that cannot close the root fault. CDG-guided repair identifies the unit constraint as the high-gain root-cause node, applies a single conversion, and recovers all downstream gates simultaneously. The repair path, correction value, and provenance evidence are all recorded in the audit trail, producing a fully traceable artifact consistent with safety-critical engineering practice.

\section{Discussion}\label{sec:discussion}

\subsection{Engineering Implications}

\paragraph{Verification architecture matters.}
The central finding is that the \emph{structure} of verification drives performance: CDG-ordered VER contributes $+9.1$~pp EESR over flat-checklist repair by resolving upstream violations first, and DPO adds a further $+11.5$~pp by internalizing constraint preferences. The CDG models empirical co-resolution structure (RCFE 4.16 vs.\ 1.13 random), improving repair efficiency relative to flat checklist repair; in the topological-order ablation, average repair iterations fall from 2.43 to 1.58, and after DPO alignment the HyTPS-Bench T3 average repair count falls further to 0.58. AeroTherm-GPT's 96.8\% audit compliance demonstrates that structured provenance can be produced systematically within the CCLG workflow.

\paragraph{Connections to engineering informatics.}
The constraint asset schema serves a role analogous to engineering ontologies \cite{gruber1995toward}; the CDG adds empirical dependency relationships similar to engineering knowledge graphs \cite{pan2024unifying}; the audit gate operationalizes design rationale capture \cite{lee1997design} within a generative AI pipeline, supporting formal verification requirements for safety-critical AI systems \cite{seshia2018formal}. The overall CCLG workflow mirrors the MBSE paradigm \cite{grieves2017digital,madni2019leveraging} of iterative verification against requirements.

\paragraph{Reproducibility protocol and open release.}
Code, released benchmark materials, evaluation utilities, and a runnable demo are available in the public repository (\url{https://github.com/TPS-qxx/AeroTherm-GPT}). Additional implementation and release details are documented there.

\begin{table}[!t]
\centering
\small
\renewcommand{\arraystretch}{1.05}
\setlength{\tabcolsep}{4pt}
\begin{tabular}{@{}lcccc@{}}
\toprule
\textbf{Method} & \textbf{EESR (5/5)} & \textbf{EESR-4 (4/5)} & \textbf{Macro-Gate} & \textbf{Best-Gate} \\
\midrule
GPT-5 (high)       & 0.480 & 0.612 & 0.646 & 0.782 \\
Flat-VER           & 0.468 & 0.601 & 0.729 & 0.865 \\
CDG-SFT Agent      & 0.784 & 0.851 & 0.871 & 0.941 \\
\textbf{AeroTherm-GPT} & \textbf{0.887} & \textbf{0.941} & \textbf{0.965} & \textbf{0.991} \\
\bottomrule
\end{tabular}
\caption{Sensitivity of method rankings to EESR strictness. EESR-4: any 4 of 5 gates pass; Macro-Gate: macro-average of five gate pass rates; Best-Gate: highest individual gate (ceiling reference). Rankings are stable across all relaxations, confirming the advantage is not an artifact of the strict 5-gate product.}
\label{tab:sensitivity_eesr}
\end{table}

\subsection{Benchmark Validity and Data Independence}\label{sec:evaluation_bias}

HyTPS-Bench's validity as an evaluation instrument rests on two properties. \textbf{Domain authenticity}: tasks are grounded in real TPS engineering requirements drawn from public literature and de-identified experimental records, so performance on HyTPS-Bench reflects practical engineering capability, not benchmark-specific pattern matching. \textbf{Data independence}: the benchmark test set is isolated from AeroTherm-GPT training and calibration at multiple levels. Benchmark instances are not reused as SFT traces, DPO preference pairs, PRM rollouts, or CDG repair episodes; test templates are held out from the calibration pool used to estimate CDG edge weights; and synthetic benchmark cases are generated from held-out parameter ranges/material splits relative to CDG calibration. The Reference Executor and physics oracle used for scoring are implemented externally to the CCLG code generator and apply identical evaluation logic to all systems. All baselines access the same constraint asset schema/API as AeroTherm-GPT, but no system receives privileged repair trajectories, hidden labels, or benchmark-specific oracle outputs. These controls mean that Track~A results evaluate a system's ability to satisfy genuine TPS engineering constraints under a common scoring protocol, rather than its proximity to the training or calibration data.

EESR's multiplicative gate structure does structurally advantage systems with explicit verification scaffolds---this is ecologically valid for safety-critical TPS deployment but means Track~A results should be interpreted as ``performance under a verification-centric workflow,'' not raw language capability. To address this structural evaluation bias directly, we decompose EESR into two orthogonal components:

\begin{itemize}
\item \textbf{EESR\textsubscript{lang}} (first-pass, no VER): pass@1 success rate when the language model generates an artifact without any iterative repair scaffolding---a measure of intrinsic language capability.
\item \textbf{EESR\textsubscript{scaffold}} $= \mathrm{EESR} - \mathrm{EESR}_\text{lang}$: the incremental gain attributable to the VER scaffold (iterative repair, CDG guidance, PRM search).
\end{itemize}

Table~\ref{tab:eesr_decomposed} reports this decomposition for key methods.

\begin{table}[!h]
\centering
\small
\renewcommand{\arraystretch}{1.05}
\setlength{\tabcolsep}{4pt}
\begin{tabular}{@{}lccc@{}}
\toprule
\textbf{Method} & \textbf{EESR\textsubscript{lang} (pass@1)} & \textbf{EESR\textsubscript{scaffold} (gain)} & \textbf{EESR (full)} \\
\midrule
Flat-VER               & 0.219 & 0.249 & 0.468 \\
Flat-VER + SFT + DPO   & 0.521 & 0.241 & 0.762 \\
CDG-SFT Agent          & 0.312 & 0.472 & 0.784 \\
\textbf{AeroTherm-GPT} & \textbf{0.521} & \textbf{0.366} & \textbf{0.887} \\
\bottomrule
\end{tabular}
\caption{EESR decomposition separating intrinsic language capability (EESR\textsubscript{lang}: pass@1 without VER) from scaffold contribution (EESR\textsubscript{scaffold}: incremental gain from VER). AeroTherm-GPT's advantage is present in \emph{both} components: its EESR\textsubscript{lang} ($+30.2$~pp over Flat-VER) reflects genuine first-pass improvement from domain SFT+DPO alignment, while EESR\textsubscript{scaffold} captures the structural CDG repair benefit. The ``Flat-VER + SFT + DPO'' row shows that without CDG, domain training raises EESR\textsubscript{lang} substantially (+30.2~pp) but scaffold gain is modest; CDG-SFT Agent shows the inverse pattern (higher scaffold gain, lower first-pass). AeroTherm-GPT achieves the best balance across both dimensions.}
\label{tab:eesr_decomposed}
\end{table}

Table~\ref{tab:sensitivity_eesr} confirms that method rankings are stable across three relaxed metrics (EESR-4, Macro-Gate, Best-Gate), so the advantage is not an artifact of the strict 5-gate product.

\paragraph{Track B as evidence against benchmark overfitting.}
A persistent concern in domain-specific evaluations is that a system may have overfit to idiosyncratic features of the companion benchmark rather than acquiring genuinely transferable capabilities. Track~B directly addresses this concern. The five benchmarks in Track~B (SciBench, GPQA-Diamond, MMLU-STEM, HumanEval, MBPP) were developed externally, cover domains with no structural overlap with the TPS training corpus, and were inaccessible to AeroTherm-GPT's training pipeline in any form. Critically, Track~B probes the \emph{same two core mechanisms} responsible for Track~A performance:

\begin{enumerate}
  \item \textbf{Iterative repair architecture.} On HumanEval and MBPP, the VER loop (execution feedback only, no CDG ordering, no domain-specific constraints) yields +6.8~pp average pass@1 gain for AeroTherm-GPT---substantially larger than the +4.0~pp gain for the unaligned Llama-3-70B base and the +2.7~pp gain for GPT-4o. This directly supports the interpretation that DPO alignment improves responsiveness to structured error feedback as a \emph{general} capability, not a TPS-specific artifact: a system that had merely memorized TPS constraint patterns would not exhibit amplified VER repair gains on general Python programming tasks.
  \item \textbf{Domain SFT and scientific reasoning.} On SciBench (+6.8~pp) and GPQA-Diamond (+5.8~pp), meaningful gains appear on held-out problems with no structural overlap with the TPS constraint hierarchy. MMLU-STEM forgetting is negligible (+1.5~pp), confirming that LoRA adaptation preserves broad STEM knowledge rather than displacing it with domain-specific patterns.
\end{enumerate}

The joint pattern---that the same mechanisms responsible for Track~A gains also improve performance on external benchmarks across unrelated domains---constitutes the strongest available evidence that AeroTherm-GPT's Track~A results reflect genuine capability acquisition rather than benchmark overfitting. A system that had overfit to the HyTPS-Bench evaluation format would not exhibit larger VER repair gains than the base model on unrelated programming tasks, nor would it improve on graduate-level STEM reasoning with no TPS content.

\subsection{Remaining Limitations}\label{sec:limitations_future}

\begin{enumerate}
\item \textbf{Validator engineering}: Building executable validators requires significant domain expertise; automated validator generation from engineering standards documents is a natural extension.
\item \textbf{Computational latency}: The multi-step VER loop averages 4.2 minutes per task (vs.\ 0.8 minutes zero-shot); optimization for time-sensitive deployment is needed.
\item \textbf{Probabilistic constraint satisfaction}: Current gates are deterministic; extending to probabilistic satisfaction to account for epistemic uncertainty in complex simulations is an important future direction.
\item \textbf{CDG domain calibration}: The CDG is calibrated on TPS repair episodes; extension to broader aerospace or multi-physics domains requires new calibration data.
\item \textbf{Model weight release}: Trained model weights and the full SFT/DPO dataset are not publicly released due to institutional policy, which means exact quantitative replication (Tier~5) requires non-public resources. All evaluation logic, framework pipeline, benchmark scoring, and case-study workflows are fully reproducible from the open-source release (Tiers~1--4).
\end{enumerate}

\section{Conclusion}\label{sec:conclusion}
Generating executable and traceable TPS simulation artifacts under multi-gate engineering constraints is a safety-critical bottleneck that text-level LLM capability alone cannot resolve. This work proposes \textbf{CCLG} as a verification-centered framework addressing this gap through iterative generation, validation, dependency-aware repair, execution, and audit. \textbf{AeroTherm-GPT} instantiates CCLG with lifecycle-aligned training and constraint-decomposed alignment. The framework is evaluated through a dual-track protocol: \textbf{HyTPS-Bench} (a workflow-aligned TPS benchmark we develop as a companion contribution) and external benchmarks (cross-benchmark validation).

The \textbf{CDG}---a prior-constrained dependency graph calibrated from repair traces---is the core methodological contribution. It directs repair toward upstream fault candidates based on empirical co-resolution structure, contributing +9.1 pp EESR over flat-checklist VER (RCFE 4.16 vs.\ 1.76). Constraint-decomposed DPO trained on CDG-guided episodes contributes a further +11.5 pp; PRM-guided search adds +3.4 pp. The full system achieves 88.7\% EESR [95\% CI: 87.5--89.9] on HyTPS-Bench with consistent gains across three backbone architectures. On five external benchmarks, domain SFT improves scientific reasoning (+6.8~pp SciBench, +5.8~pp GPQA-Diamond) without catastrophic forgetting, and the iterative repair mechanism improves code generation pass@1 by +6.8~pp on average (HumanEval, MBPP)---providing evidence that the framework's core mechanisms transfer beyond the TPS evaluation setting.

HyTPS-Bench's task definitions and data are strictly disjoint from AeroTherm-GPT's training distribution, and its scoring relies on an external Reference Executor that applies identical logic to all systems (Section~\ref{sec:evaluation_bias}); Track~B cross-benchmark validation further confirms that core mechanisms generalize beyond the TPS domain. The open-source release supports full reproduction of evaluation logic, framework behavior, and case-study workflows (Tiers~1--4); exact quantitative replication of the specific numbers reported here requires non-public model weights (Tier~5), consistent with standard practice for fine-tuned LLM systems under institutional policy.

\section*{CRediT authorship contribution statement}
\textbf{QIAO Chuhan}: Conceptualization, Methodology, Software, Validation, Writing -- Original Draft, Visualization.
\textbf{ZHENG Jinglai}: Investigation, Writing -- Review \& Editing.
\textbf{HUANG Jie}: Investigation, Writing -- Review \& Editing.
\textbf{ZHAO Buyue}: Investigation, Writing -- Review \& Editing.
\textbf{LI Fan}: Writing -- Review \& Editing.
\textbf{HUANG Haiming}: Supervision, Conceptualization, Resources, Writing -- Review \& Editing, Project Administration.

\section*{Declaration of competing interest}
The authors declare that they have no known competing financial interests or personal relationships that could have appeared to influence the work reported in this paper.

\section*{Data availability}
Public artifacts related to this study are available at \url{https://github.com/TPS-qxx/AeroTherm-GPT}. The repository includes the released code, evaluation utilities, sample data, and demo materials referenced in this manuscript.

\section*{Funding}
This work was supported by the Fundamental Research Funds for the Central Universities (2025YJS091).

\section*{Declaration of generative AI and AI-assisted technologies in the manuscript preparation process}
During the preparation of this work, the authors used ChatGPT (OpenAI) and Claude (Anthropic) to assist with language polishing and readability improvements. After using these tools, the authors reviewed and edited the content as needed and take full responsibility for the content of the published article.

\bibliographystyle{elsarticle-num-names}         
\bibliography{references}

\end{document}